\documentclass{article}
\PassOptionsToPackage{numbers, compress}{natbib}
\usepackage[dblblindworkshop, final]{tagds_2025}
\workshoptitle{Topology, Algebra, and Geometry in Data Science}
 
\usepackage[utf8]{inputenc} % allow utf-8 input
\usepackage[T1]{fontenc}    % use 8-bit T1 fonts
\usepackage{hyperref}       % hyperlinks
\usepackage{url}            % simple URL typesetting
\usepackage{booktabs}       % professional-quality tables
\usepackage{amsfonts}       % blackboard math symbols
\usepackage{amsthm}         % proof environment
\usepackage{nicefrac}       % compact symbols for 1/2, etc.
\usepackage{microtype}      % microtypography
\usepackage{xcolor}         % colors
\usepackage{comment}
\usepackage{amsmath}

\usepackage{amsmath}
\usepackage{graphicx} 
\usepackage{subcaption}
\usepackage{algorithm}         % floating “Algorithm” environment
\usepackage{algorithmic}       % “algorithmic” environment (for \Require, \Ensure, \State, etc.)

\usepackage{tikz}
\usetikzlibrary{calc}
\usepackage{wrapfig}

\definecolor{naivec}{RGB}{192,62,53}
\definecolor{latentc}{RGB}{203, 97, 32}
\definecolor{scalegmn}{RGB}{190,146,7}
\definecolor{rebasin}{RGB}{47,150,141}
\definecolor{ngraphs}{RGB}{115,94,181}

\newtheorem{proposition}{Proposition}

\title{Symmetry-Aware Graph Metanetwork Autoencoders: Model Merging through Parameter Canonicalization}

\author{%
    Odysseas Boufalis$^1$\footnotemark[1] \qquad
    Jorge Carrasco-Pollo$^1$\footnotemark[1] \qquad
    Joshua Rosenthal$^1$\footnotemark[1] \qquad \\
    \vspace{0.5em}
    \textbf{Eduardo Terrés-Caballero}$^1$\footnotemark[1] \qquad
    \textbf{Alejandro García-Castellanos}$^2$ \\
    $^1$Informatics Institute, University of Amsterdam \\
    $^2$Amsterdam Machine Learning Lab (AMLab), University of Amsterdam
}
\begin{document}

\maketitle
\renewcommand{\thefootnote}{\fnsymbol{footnote}}\footnotetext[1]{
    Authors ordered alphabetically. Equal contribution.
    Correspondence: <\texttt{a.garciacastellanos@uva.nl}>
}

\footnotetext[2]{\url{https://github.com/odyboufalaki/Symmetry-Aware-Graph-Metanetwork-Autoencoders}}

\begin{abstract}
  Neural network parameterizations exhibit inherent symmetries that yield multiple equivalent minima within the loss landscape. Scale Graph Metanetworks (ScaleGMNs) explicitly leverage these symmetries by proposing an architecture equivariant to both permutation and parameter scaling transformations. Previous work by Ainsworth et al.~(2023) addressed permutation symmetries through a computationally intensive combinatorial assignment problem, demonstrating that leveraging permutation symmetries alone can map networks into a shared loss basin. In this work, we extend their approach by also incorporating scaling symmetries, presenting an autoencoder framework utilizing ScaleGMNs as invariant encoders. Experimental results demonstrate that our method aligns Implicit Neural Representations (INRs) and Convolutional Neural Networks (CNNs) under both permutation and scaling symmetries without explicitly solving the assignment problem. This approach ensures that similar networks naturally converge within the same basin, facilitating model merging, i.e., smooth linear interpolation while avoiding regions of high loss. The code is publicly available on our GitHub repository\footnotemark[2].
\end{abstract}

\section{Introduction}

Model merging has emerged as a powerful technique for combining the capabilities of independently trained neural networks, originally introduced to address Linear Mode Connectivity (LMC) barriers~\cite{garipov_2018}. LMC barriers refer to the obstacles encountered when attempting to linearly interpolate between independently trained networks, which often results in suboptimal performance~\cite{ruan2025task}. However, a fundamental challenge arises from the fact that neural networks exhibit extensive \textit{parameter-space symmetries}: systematic permutations and scalings of parameters that leave the network's function unchanged \cite{zhao2025symmetry}. These symmetries create multiple equivalent representations of the same function scattered across different regions of parameter space, making direct interpolation between network weights ineffective and often resulting in high loss barriers that hinder linear mode connectivity.

The standard solution is to \emph{explicitly align} networks before merging by finding correspondences between their parameters. One approach uses numerical solvers: Git Re-Basin \cite{Ainsworth_2023_GitReBasin} formulates alignment as a layer-wise linear assignment problem solved via the Hungarian algorithm \cite{Kuhn_1955_Hungarian}, but this scales poorly with network size due to combinatorial complexity and typically operates only on pairs of networks. An alternative approach employs \emph{metanetworks}, i.e., neural networks that take other networks as input, to directly learn alignment mappings \cite{Gibbons_2019, Lee_2018, Kim_2021, navon2023equivariant}. These methods attempt to approximate optimal permutations of weights through learned transformations, bypassing combinatorial solvers.

However, current approaches face significant limitations when handling the full spectrum of parameter symmetries. While some existing methods do address scaling symmetries \cite{Zhang_2025_Transformers, pittorino2022deep}, they either operate only on specific architectures or employ heuristic approaches that sequentially handle different transformation types—first solving for one type of weight transformation (typically permutations) and then addressing scalings as a separate step. As we demonstrate in the Appendix (Section~\ref{appendix:rebasin-generalization}), these sequential heuristics can be effective for simple cases like sign flip symmetries but may prove adversarial when dealing with general scaling symmetries, potentially degrading rather than improving alignment quality.

This limitation is particularly significant given recent evidence that scaling symmetries play a crucial role in understanding neural networks' weight space \cite{zhao2025symmetry, zhao2022symmetries,lengyel2020genni, Zhao_2023_Connectivity, garcia2024relative}. Indeed, it has been shown that acknowledging scaling symmetries substantially improves training dynamics during optimization~\cite{zhao2025symmetry}. However, there has been no systematic analysis of how general scaling symmetries interact with model merging quality, nor methods that jointly handle both permutation and scaling symmetries in a principled, architecture-agnostic manner—gaps that motivate our work.

These limitations motivate a fundamentally different approach: rather than explicitly computing alignments, we can \emph{implicitly canonicalize} networks by learning to map them into a symmetry-invariant representation space. Such an approach requires an encoder that is equivariant to the relevant parameter symmetries. Scale Graph Metanetworks (ScaleGMNs) \cite{Kalogeropoulos_2024} provide exactly this property through their graph-based neural network representation that is provably equivariant to both permutations and scalings.

Building on this foundation, we propose a canonicalization autoencoder framework where a ScaleGMN encoder maps neural networks into symmetry-invariant latent codes, and an MLP decoder reconstructs canonical parameter representations from these codes. This approach automatically handles both permutation and scaling symmetries without combinatorial optimization, scales efficiently with network size, and generalizes across multiple networks simultaneously. The resulting canonical representations enable effective linear interpolation and provide new insights into the geometry of neural network parameter spaces.

In summary, our main contributions are:
% \begin{itemize}
%     \item A symmetry-aware autoencoding framework that implicitly canonicalizes neural networks by collapsing both permutation and scaling orbits in parameter space.
%     \item Comprehensive analysis on the impact of scaling symmetries on model interpolation quality.

%     \item To enable a fair comparison, we generalize the Git Re-Basin algorithm to also handle sign-flip symmetries.
    
%     \item Empirical validation on implicit neural representations and CNNs demonstrating superior linear mode connectivity compared to existing explicit alignment methods.
% \end{itemize}

\begin{itemize}
    \item A novel symmetry-aware autoencoding framework that achieves implicit neural network canonicalization by jointly collapsing permutation and scaling symmetry orbits, eliminating the need for explicit alignment procedures.
    
     \item Comprehensive analysis on the impact of scaling symmetries on model interpolation quality.
    
    \item An extended Git Re-Basin algorithm that incorporates sign-flip symmetries, establishing a stronger baseline for symmetry-aware neural network alignment.
    
    \item Extensive experimental validation across implicit neural representations and CNNs, demonstrating significant improvements in linear mode connectivity over existing explicit alignment methods.
\end{itemize}

\section{Related Work}
\label{sec:rel}

\paragraph{Weight-space alignment.}
  Model merging seeks to combine multiple trained models into one. In an ideal case, if models reside within the same loss basin, they possess linear mode connectivity and can be merged through a simple linear interpolation of their weights while maintaining low loss \cite{garipov_2018,Lin_2024_ExploringNeuralNetworkLandscapes}. However, this ideal condition is rarely met in practice, mainly because of the inherent symmetries of neural networks, such as permutation symmetry, which states that reordering the neurons within a hidden layer (and its connected weights) does not alter the overall function of the network \cite{Hecht_1990}.

 Because of this symmetry, models trained with different random initializations tend to converge to functionally equivalent but geometrically distinct solutions, preventing a naive interpolation from succeeding~\cite{Ainsworth_2023_GitReBasin,Stoica_2024_ZipIt}. The key insight of modern merging techniques is to exploit this very property. \citet{Ainsworth_2023_GitReBasin} do this by finding an optimal permutation $P_i$ to align the neurons of one model with another by solving the \textit{assignment problem}, thus bringing the models into a shared basin. Once aligned, the simple linear interpolation can be successfully applied to the weights of a given layer $i$:
\[
W_i^*=\gamma W_i^A+(1-\gamma)P_iW_i^BP^\top_{i-1}
\]
 with $\gamma \in[0,1]$, where $P_i$ permutes the output dimension of layer $i$ and $P_{i-1}$ permutes its input dimension.

   However, solving the assignment problem is NP-hard, and therefore usually requires using heuristic iterative optimization algorithms such as Git Re-Basin \cite{Ainsworth_2023_GitReBasin}. On the other hand, an alternative paradigm uses a dedicated neural network for amortizing this task \citep{Gibbons_2019, Lee_2018, Kim_2021}. Among these, \textsc{Deep-Align}~\citep{navon2023equivariant} trains a permutation-equivariant architecture to predict the layer-wise permutation matrices in a single forward pass. By learning the alignment task itself, \textsc{Deep-Align}  directly outputs the optimal transformation, bypassing the need for separate, iterative solvers. 
   
   % ZipIt! \citep{Stoica_2024_ZipIt} generalizes neuron alignment beyond strict across-model permutations. It allows within-model matches and partial layer merging, enabling effective parameter alignment even across tasks with low feature similarity.   

    As an alternative to explicit alignment, \citet{Lim_2024_Empirical} show that reducing parameter symmetries in the network architecture itself can render permutation-based alignment for model merging unnecessary. 
    
    Similarly, our method yields canonical network representations without explicitly solving combinatorial alignment problems, relying instead on symmetry-aware encoder backbones combined with an appropriate functional loss. Beyond permutations, neural networks also exhibit scaling symmetries related to activation functions \cite{Godfrey_2023_Symmetries}. Our work seeks to account for both symmetries.

    \paragraph{Weight-space networks and latent representations.}    
    Metanetworks \citep{zhou2023permutation, zhou2024universal, Navon_2023_EquivariantArchitectures, tran2024monomial, Zhou_2023_NeuralFunctionalTransformers} treat each data point as a distinct neural network, with symmetries extensively studied in these architectures. Graph Metanetworks (GMNs) \citep{Lim_2023_GMN, Kofinas_2024_Graph} represent neural networks as graphs, using permutation-equivariant graph neural networks (GNNs) for processing. ScaleGMNs \citep{Kalogeropoulos_2024} extend GMNs by incorporating scaling symmetries for MLPs and CNNs, demonstrating enhanced performance on tasks such as INR classification, network editing, and generalization prediction.
    
    Another application of metanetworks is embedding model weights into meaningful latent representations. \citet{DeLuigi_2023} pioneered this approach using a simple MLP to map vectorized weights of Neural Fields to latent vectors. The authors showed that these latent vectors encode both the semantics and properties of the original network, including performance characteristics. Building on this work, \citet{Zhou_2023_NeuralFunctionalTransformers} presented \textsc{Inr2Array}, an autoencoder that maps model weights into a latent point cloud and reconstructs multiple weights representing local patches in the input domain of the original Neural Field.
    
    Our method also constructs an autoencoder using an invariant encoder, but differs significantly from \textsc{Inr2Array} in two key aspects:
    
    \begin{itemize}
        \item \textbf{Encoder architecture:} While \citet{Zhou_2023_NeuralFunctionalTransformers} employ Neural Functional Transformers (NFTs) as their invariant encoder, we propose using Graph Metanetworks such as ScaleGMNs. Although NFTs offer high scalability, they require weight vectorization during the attention mechanism and do not actively leverage the computational graph structure as ScaleGMNs do. Additionally, NFTs are only equivariant with respect to permutations, whereas our approach acknowledges the full symmetry group of the network.
    
        \item \textbf{Latent representation and decoding:} The latent point cloud representation and decoding of multiple subnetwork weights can provide an expressive latent code for INRs \cite{bauer2023spatial}. However, this configuration limits their applications. We propose using a single invariant latent vector and decoding a single canonical weight configuration. This approach enables us to encode networks beyond INRs, such as CNN classifiers, and utilize the canonicalized weights for model merging applications.
    \end{itemize}
    
    % Our approach leverages the ScaleGMN as the backbone of our autoencoder. 
    
    % Autoencoders \cite{Rumelhart_1986_Autoencoder} map inputs to meaningful representations and reconstruct the original data. \citet{DeLuigi_2023} applied autoencoding effectively to INRs. Our work adapts this setup using ScaleGMNs to encode INRs into invariant representations. Decoding these representations yields canonical neural network forms, facilitating interpolation without solving explicit assignment problems.

\section{Preliminaries}
    \subsection{Implicit Neural Representations}

    Implicit Neural Representations (INRs), also known as Neural Fields, encode signals such as images, audio, or 3D shapes directly within neural network parameters, mapping continuous input coordinates to corresponding output values learned from discrete samples \citep{DeLuigi_2023}. While standard MLP-based INRs are resolution-independent, they often struggle to capture high-frequency details due to spectral bias, leading to overly smooth reconstructions of signals with fine structures. The sinusoidal representation network (SIREN) architecture \citep{Sitzmann_2020_Siren} addresses this limitation by replacing conventional activations with periodic sine functions, enabling faithful representation of both low and high frequency components and improving convergence in tasks like neural rendering and physical simulation. Since INRs store the entirety of the signal information in their weights, architectures processing these representations must also account for inherent parameter-space symmetries that can give rise to redundant minima~\cite{Hecht_1990}.
    
    % \subsection{Convolutional Neural Networks}

    \subsection{Permutation and scaling symmetries}
        In this section, we review recent work that investigates the inherent symmetries in neural networks~\cite{Godfrey_2023_Symmetries}. The key insight is that certain activation functions create hidden symmetries in the weight space, under which networks with different parameter values can exhibit functionally equivalent behavior.
        
        Specifically, they prove that for many activation functions $\sigma$, the \textit{intertwiner group} $I_{\sigma,d}$ (matrices $A$ such that $\sigma(Ax) = B\sigma(x)$) consists of matrices of the form $PQ$. Here, $P$ is a permutation matrix, and $Q$ is a diagonal matrix $\text{diag}(q_1, \dots, q_d)$ where $q_i$ belong to $D_{\sigma}$, a 1-dimensional group specific to $\sigma$.

        For ReLU, Proposition 3.4 in \cite{tran2024monomial} states that the intertwiner matrices $A = P Q$ are \textbf{generalised permutation matrices with positive entries}, i.e., $Q = \mathrm{diag}(q_1, \dots, q_d)$ with $q_i > 0$. For SIRENs, which use the sine activation $\sigma(x) = \sin(\omega x)$, as well as for the $\tanh$ activation function, the same proposition specifies that $A$ is a \textbf{signed permutation matrix}, meaning the entries $q_i$ are restricted to $\pm 1$. Moreover, for all the aforementioned activations, the transformation $B$ coincides with $A$.
        
  % Consequently, the scaling component $Q$ within these symmetries depends on the activation function: 
  %   \begin{itemize}
  %       \item for ReLU, $Q$ has strictly positive diagonal entries (positive scalings),
  %       \item for sine and $\tanh$, $Q$ is restricted to \textbf{sign flips} (diagonal entries $\pm 1$).
  %   \end{itemize}
Applied layer-wise in a network, these symmetries take the form
\[
W'_{\ell} = (Q_{\ell} P_{\ell}) W_{\ell} (P_{\ell-1}^{\top} Q^{-1}_{\ell-1}), 
\quad
b'_{\ell} = (Q_{\ell} P_{\ell}) b_{\ell}
\quad\Longrightarrow\quad u_{\theta'}(x) = u_{\theta}(x).
\]
The $(Q_{\ell}P_{\ell})$ factor reorders the neurons in layer $\ell$ and rescales them---by positive factors for ReLU, or by $\pm 1$ for sine/$\tanh$. 
To preserve the network’s output, $W'_{\ell}$ is transformed by $(Q_{\ell}P_{\ell})$ on its output side, while $(P_{\ell-1}^{\top}Q^{-1}_{\ell-1})$ adjusts the input side to remain compatible with the preceding layer’s transformation.

To address these symmetries, \cite{Kalogeropoulos_2024} introduce ScaleGMN, a specialized type of GMN that represents neural networks as computational graphs, with neurons as nodes and connections as edges (see Appendix~\ref{appendix:graph_metanetworks} for details).
\section{Neural Network Autoencoding for Canonicalization}

We propose an autoencoder framework for canonicalizing neural networks in parameter space with respect to a predefined set of symmetries. The encoder maps input parameters into a symmetry-invariant latent vector, while a simple MLP decoder reconstructs a canonical representative of the network. The encoder's architectural invariance ensures a unique latent representation for symmetry-equivalent networks. This canonicalization technique aligns with recent work in Geometric Deep Learning such as \cite{kaba2023equivariance}, which learns a representative for each orbit by constructing a $G$-invariant function $f_{\theta} : X \to X$ such that $f_{\theta}(x) \in Gx$ for each $x \in X$. This approach contrasts with moving frame techniques \cite{olver2001joint}, which instead learn the group action that maps elements to their orbit representative. As discussed in Section~\ref{sec:rel}, the \textsc{Deep-Align} model \citep{navon2023equivariant} belongs to this latter category.

\begin{figure}[ht!]
    \centering
    \includegraphics[width=0.9\linewidth]{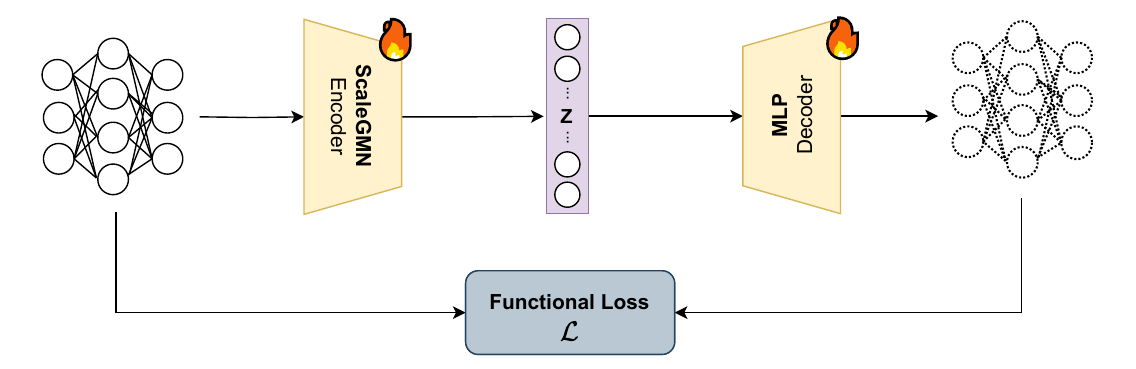}
    \caption{Autoencoder architecture for neural network canonicalization using a permutation and scaling-invariant ScaleGMN encoder, MLP decoder, and functional loss to preserve network equivalence.}
    \label{fig:autoencoder-architecture}
\end{figure}

Our experiments employ two symmetry-aware encoders: Neural Graphs \cite{Kofinas_2024_Graph}, which is permutation-invariant, and ScaleGMN \cite{Kalogeropoulos_2024}, which is both permutation and scale-invariant. This choice lets us disentangle the effect of permutation invariance and measure the added value of scale invariance. The architectural differences between these encoders also highlight the robustness of our framework across diverse designs. Figure~\ref{fig:autoencoder-architecture} illustrates the pipeline using ScaleGMN.
More scalable permutation-invariant metanetworks such as UNFs \cite{zhou2024universalneuralfunctionals} could extend this approach to larger architectures, including Transformers. Since ScaleGMN does not yet support such architectures, we focused on testing the autoencoder’s generalizability across symmetries rather than model size. Nonetheless, scaling to larger networks is straightforward given an appropriate encoder.

% We use ScaleGMNs—one of the most capable scale-equivariant models—specifically to analyze the role of scaling symmetries in model merging.
% More scalable metanetworks such as UNFs\cite{zhou2024universalneuralfunctionals}, which are only permutation-equivariant \textcolor{red}{(shouldnt we say invariant?, since we dont really care bout equivariance here)}, could extend this approach to larger architectures, including transformers.
% However, the original study on ScaleGMN does not provide this extension to transformers \textcolor{red}{(even though they mention it in future work ? double check)} hence we prioritize studying the generalizability of the autoencoder pipeline to diverse
% multiple symmetries to studying bigger architectures like transformers. However this is not a limitation and the pipeline here presented can trivially be extended by using the adequate encoder.

% A justification should have been added in the intro for why scale and permutation are important

% Using ScaleGMN as the encoder, we process networks in a manner invariant to permutations and scalings, thereby producing a canonical latent representation.

    % We apply this pipeline to INRs and CNNs. The goal is for the decoded network to be functionally equivalent to the input network while mapping outputs into a single basin. To achieve this, we use a functional loss that enforces equivalence directly. For INRs, we regress pixel activations of the image represented by the INR; for CNN classifiers, we minimize the KL divergence between the probability distributions derived from each network’s logits via a temperature-scaled softmax.

    % Moreover, we investigate the importance of scaling symmetries by also training an autoencoder whose encoder is a Neural Graphs network \cite{Kofinas_2024_Graph}, which does not account for scaling symmetries.

    We apply this pipeline to INRs and CNNs. The objective is for the decoded network to be functionally equivalent to the input while mapping outputs into a single basin. To enforce this, we use a functional training loss: for INRs, we regress the pixel activations of the image represented by the INR; for CNN classifiers, we minimize the KL divergence between probability distributions obtained from each network’s logits via a temperature-scaled softmax. Notably, this loss formulation could be directly applied to canonicalizing other architectures with probabilistic outputs, such as Transformers, where the autoregressive objective relies on a softmax over the vocabulary. For CNNs we consider two activation variants—ReLU and \textit{tanh}—to probe symmetry effects: ReLU induces positive-scaling symmetries across adjacent layers, whereas \textit{tanh} only induces sign-flip symmetry. To assess the role of these symmetries, we also train an autoencoder whose encoder is a Neural Graphs network \cite{Kofinas_2024_Graph}, which does not account for scaling.

\section{Experiments and results}
\label{sec:exp}

In this section, we empirically validate our autoencoder-based canonicalization framework. We evaluate its ability to establish high-performance linear paths between neural networks, a key indicator of successful model merging. All experiments are performed on a single H100 GPU. %Our experiments span two distinct settings: interpolating functionally equivalent Implicit Neural Representations (INRs) and functionally \textcolor{red}{distinct(?)} Convolutional Neural Networks (CNNs).

\subsection{Experimental setup}

\paragraph{Tasks and Datasets.} We evaluate our canonicalization approach across two complementary scenarios that test different aspects of network alignment:

% \textit{(1) Aligning INRs:} We test the ability to connect functionally \textit{identical} but parametrically distinct networks. The networks are INRs (MLPs with sine activations) \cite{Navon_2023_EquivariantArchitectures} trained to represent MNIST~\cite{lecun-mnisthandwrittendigit-2010} digits. To obtain functionally equivalent endpoints, we start from a trained INR, apply a series of group actions that leave the function unchanged, and add small noise to prevent them from mapping to the same latent point, since the ScaleGMN encoder is invariant to the applied group actions. We report interpolation results without resulting to adding noise on Appendix (Section~\ref{appendix:interpolation-without-noise}). The goal is to recover a zero-barrier interpolation path. Moreover, we vary the group actions used to construct the interpolation endpoints to be drawn solely from permutation symmetries, solely from scaling symmetries, or from both. To enable a fair comparison, we generalize the Git Re-Basin algorithm to also handle scaling symmetries; the details of this generalization are provided in Appendix (Section~\ref{appendix:rebasin-generalization}).

\textit{(1) Aligning INRs:} We test alignment between functionally \textit{identical} but parametrically distinct Implicit Neural Representations (INRs). These networks are MLPs with sine activations trained to represent MNIST digits~\cite{lecun-mnisthandwrittendigit-2010} introduced in ~\cite{Navon_2023_EquivariantArchitectures}. To create functionally equivalent endpoints, we start from a trained INR, apply intertwiner group transformations (permutations and/or scaling), and add small perturbations to prevent identical latent mappings. We systematically vary the applied symmetries (using permutations only, scaling only, or both) to isolate the contribution of different symmetry types. The goal is to recover perfect linear mode connectivity with zero interpolation barrier. To enable a fair comparison, we generalize the Git Re-Basin algorithm to also handle scaling symmetries; the details of this generalization are provided in the Appendix (Section~\ref{appendix:rebasin-generalization}). Also, as an additional ablation, we report interpolation results without resorting to adding noise in the Appendix (Section~\ref{appendix:interpolation-without-noise}).

\textit{(2) Aligning CNN Classifiers:} We evaluate the ability to connect pairs of CNNs that are not necessarily functionally equivalent. We interpolate between pairs of CNNs with distinct CIFAR-10 test accuracies, selected from the SmallCNN Zoo dataset~\cite{unterthiner2021predictingneuralnetworkaccuracy}. We report interpolation results for randomly sampled CNN pairs as well as averaged results over 20 pairs, which are drawn from the top 1,500 highest-performing models in the SmallCNN Zoo. The models are simple CNNs with a [16, 16, 16] hidden channel architecture, 3x3 convolutions, and an average pooling layer. The goal is to find a high-performance path that avoids the typical accuracy drop.

\paragraph{Compared Methods.}
We benchmark our proposed method against several baselines:
\begin{itemize}
    \item \textbf{Naive Interpolation:} A standard baseline involving linear interpolation in the original parameter space.
    \item \textbf{Linear assignment} (w/ Git Re-Basin~\cite{Ainsworth_2023_GitReBasin}): An optimization-based method that explicitly solves the layer-wise assignment problem to align permutation symmetries. As previously mentioned, for the INR experiments, we use our generalized version (Appendix, Section~\ref{appendix:rebasin-generalization}) that also handles sign-flip symmetries.
    \item \textbf{Neural Graphs Autoencoder (Ours):} Our autoencoder framework equipped with a permutation-only equivariant encoder~\cite{Kofinas_2024_Graph} to ablate the effect of handling scaling symmetries.
    \item \textbf{ScaleGMN Autoencoder (Ours):} Our proposed autoencoder using the ScaleGMN encoder, which is equivariant to both permutation and scaling symmetries.
\end{itemize}

\paragraph{Evaluation Metrics.}
For both tasks, we evaluate the quality of the linear interpolation path between two networks, $\theta_A$ and $\theta_B$, after alignment. For INRs (Figure~\ref{fig:interpolation-inr}), we measure the reconstruction loss at intervals along the path, where a lower barrier indicates better alignment. For CNNs (Figures~\ref{fig:interpolation-cnn-relu-multiple-pairs} and ~\ref{fig:interpolation-cnn-tanh-multiple-pairs}), we measure both the CIFAR-10 test accuracy and the cross-entropy loss along the path, where successful linear mode connectivity is indicated by a monotonic, high-accuracy curve and a correspondingly low loss barrier.

\subsection{Results}
\label{section:results}

\paragraph{Autoencoder Performance.}
We report on the reconstruction capacity of our autoencoder framework using distinct metrics for each task. For the INR setting (Table~\ref{tab:inr-reconstruction}), reconstruction quality is quantified by the mean squared error (MSE) between the pixel activations of the original and decoded networks. For the CNN setting (Table~\ref{tab:cnn-reconstruction}), we assess functional preservation via two metrics: the L1 error, representing the absolute difference in CIFAR-10 test accuracy, and Kendall’s Tau \cite{Kendall1938}, a rank correlation coefficient that quantifies how well the performance ordering of networks is maintained after reconstruction.
    
    \begin{table}[htbp]
    \vspace{-15pt}
        \centering
        \begin{minipage}{0.4\textwidth}
            \centering
            \caption{Reconstruction error of autoencoder variants for the Implicit Neural Representation (INR) task.}
            \medskip
            \label{tab:inr-reconstruction}
            \resizebox{\textwidth}{!}{
            \begin{tabular}{lcc}
                \toprule
                \textbf{Model} & \multicolumn{2}{c}{\textbf{Reconstruction Error ($\downarrow$})} \\
                \cmidrule(lr){2-3}
                               & \textbf{Train} & \textbf{Test} \\
                \midrule
                ScaleGMN       & 0.0096 & \textbf{0.0106} \\
                Neural Graphs  & \textbf{0.0068} & 0.0135 \\
                \bottomrule
            \end{tabular}
            }
        \end{minipage}
        \hfill
        \begin{minipage}{0.56\textwidth}
            \centering
            \caption{L1-Error and Kendall's Tau of autoencoder variants on the test set for the CNN canonicalization.}
            \medskip
            \label{tab:cnn-reconstruction}
            \resizebox{\textwidth}{!}{
            \begin{tabular}{lcc}
                \toprule
                \textbf{Model} & \textbf{L1-Error ($\downarrow$)} & \textbf{Kendall's Tau ($\uparrow$)} \\
                \midrule
                \multicolumn{3}{l}{\textit{ReLU Variants}} \\
                ScaleGMN-ReLU  & \textbf{0.0111} & \textbf{0.9100} \\
                Neural Graphs-ReLU & 0.0142 & 0.8914 \\
                \midrule
                \multicolumn{3}{l}{\textit{Tanh Variants}} \\
                ScaleGMN-tanh  & \textbf{0.0090} & \textbf{0.9160} \\
                Neural Graphs-tanh & 0.0113 & 0.8905 \\
                \bottomrule
            \end{tabular}
            }
        \end{minipage}
    \end{table}

    Model selection mirrors the evaluation metrics: for INRs we choose by validation MSE; for CNNs by validation L1 error, aiming to preserve the original accuracy as closely as possible. Hyperparameter tuning and other experimental details for both autoencoders are provided in Appendix ~\ref{appendix:experimental-details}.

\begin{figure}[h!]
    \centering
    \begin{subfigure}[t]{0.45\textwidth}
        \centering
        \includegraphics[width=\linewidth]{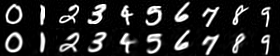}
        \caption{ScaleGMN reconstruction.}
        \label{fig:sgmn_reconstruction}
    \end{subfigure}
    \hfill
    \begin{subfigure}[t]{0.45\textwidth}
        \centering
        \includegraphics[width=\linewidth]{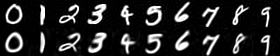}
        \caption{Neural Graphs reconstruction.}
        \label{fig:baseline_reconstruction}
    \end{subfigure}

    \caption{
        Comparison between the ground truth (top row) and reconstruction obtained with the different types of INR autoencoders (bottom row), for a set of distinct MNIST digits.
    }
    \label{fig:reconstructions}
\end{figure}

In Figure \ref{fig:reconstructions}, we visualize the output of our autoencoder across the two experimental autoencoder setups. We see that in each case, the autoencoder is able to successfully recover the input INR, producing mappings with high visual fidelity.

\paragraph{INR Interpolation.}
As shown in Figure~\ref{fig:interpolation-inr}, naive interpolation fails with a significant loss barrier, confirming the endpoints lie in separate basins. Both Git Re-Basin and our autoencoder approach succeed in establishing near-linear mode connectivity across all three transformation types (permutation, scaling, and their combination), with the autoencoder exhibiting more stable performance, as indicated by the smaller standard deviation (shaded region).

% % Legend item: band + line to mimic your plots
% \newcommand{\legendband}[2]{%
%   \begin{tikzpicture}[baseline=-0.6ex]
%     % \path[fill=#1, fill opacity=.20, draw=#1, line width=1.0pt]
%     %   (0,0.35ex) rectangle (10mm,1.7ex);
%     \draw[#1, line width=1pt] (0,0.1ex) -- (10mm,0.1ex);
%   \end{tikzpicture}\hspace{0.4em}#2%
% }

%     \begin{figure}[h!]
%   \centering
%   % --- Legend on top ---
%   \begin{minipage}{\linewidth}\centering\footnotesize
%     \legendband{naivec}{Naive}\hspace{1.2em}
%     \legendband{rebasin}{Linear Assignment}\hspace{1.2em}
%     \legendband{ngraphs}{Neural Graphs}\hspace{1.2em}
%     \legendband{scalegmn}{ScaleGMN}\hspace{1.2em}
%   \end{minipage}

%   \vspace{4pt}

%   % --- Your panels ---
%   \begin{subfigure}[b]{0.32\linewidth}
%     \includegraphics[width=\linewidth]{media/interpolation_inr/interpolation-P-perturbation.pdf}
%     \caption{Permutations}
%   \end{subfigure}
%   \begin{subfigure}[b]{0.32\linewidth}
%     \includegraphics[width=\linewidth]{media/interpolation_inr/interpolation-D-perturbation.pdf}
%     \caption{Scaling}
%   \end{subfigure}
%   \begin{subfigure}[b]{0.32\linewidth}
%     \includegraphics[width=\linewidth]{media/interpolation_inr/interpolation-PD-perturbation.pdf}
%     \caption{Permutations + Scaling}
%   \end{subfigure}

%   \caption{Interpolation experiments.}
%   \label{fig:interpolation-inr}
% \end{figure}

% Enhanced legend item: line with proper baseline alignment
% Enhanced legend item: line with proper baseline alignment
\newcommand{\legenditem}[2]{%
  \begin{tikzpicture}[baseline=-0.5ex]
    \draw[#1, line width=2.4pt, line cap=round] (0,0) -- (4mm,0);
  \end{tikzpicture}\hspace{0.5em}\textbf{#2}%
}
% Legend box with border
\newcommand{\legendbox}[1]{%
  \begin{tikzpicture}
    \node[
      draw=black!60,
      line width=0.8pt,
      rounded corners=3pt,
      fill=white,
      inner sep=4pt
    ] {
      \begin{minipage}{0.85\linewidth}
        \centering\small
        #1
      \end{minipage}
    };
  \end{tikzpicture}
}

\begin{figure}[h!]
  \centering
  
  % --- Enhanced Legend with border ---
  \legendbox{%
    \legenditem{naivec}{Naive}\hspace{1.5em}%
    \legenditem{rebasin}{Linear Assignment}\hspace{1.5em}%
    \legenditem{ngraphs}{Neural Graphs}\hspace{1.5em}%
    \legenditem{scalegmn}{ScaleGMN}%
  }
  
  \vspace{4pt}
  
  % --- Your panels ---
  \begin{subfigure}[b]{0.32\linewidth}
    \includegraphics[width=\linewidth]{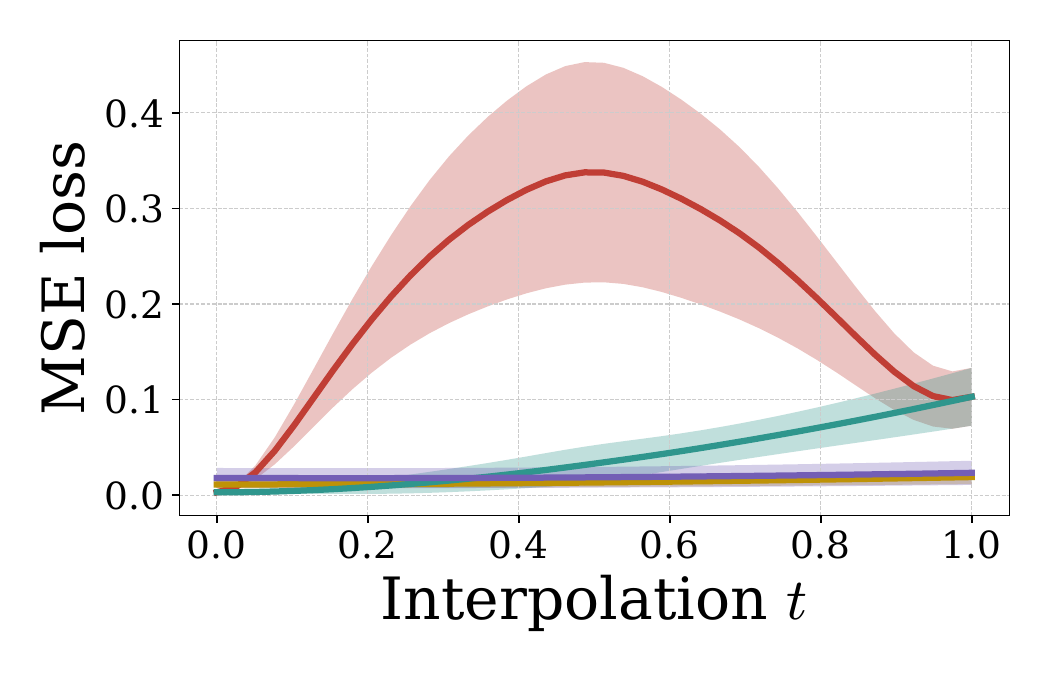}
    \caption{Permutations}
  \end{subfigure}
  \hfill\begin{subfigure}[b]{0.32\linewidth}
    \includegraphics[width=\linewidth]{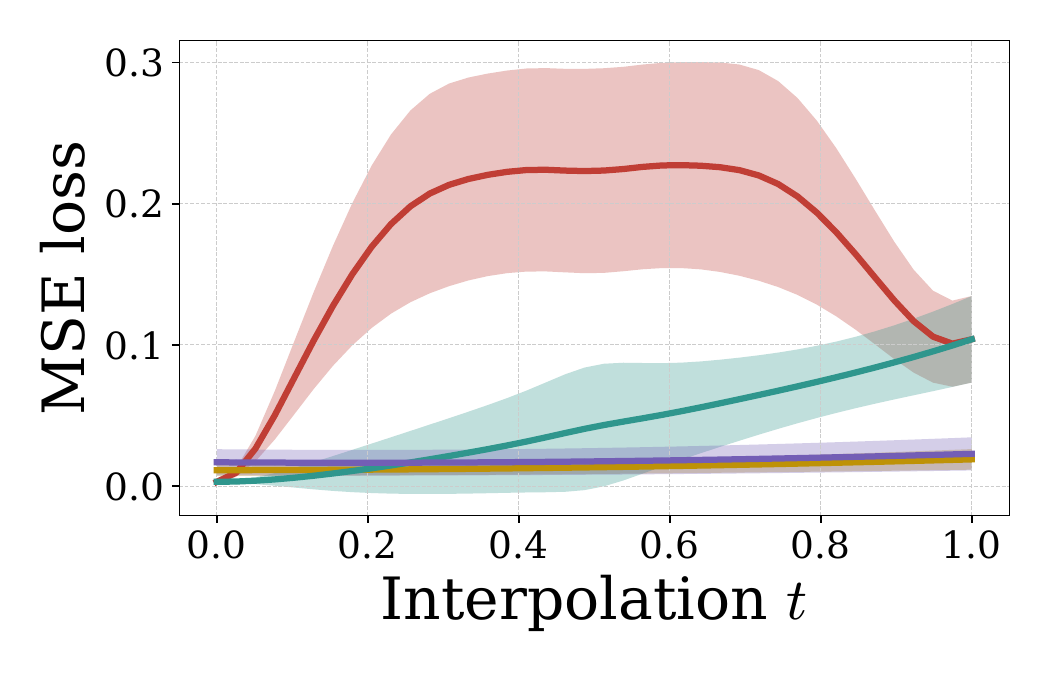}
    \caption{Scaling}
  \end{subfigure}
  \hfill
  \begin{subfigure}[b]{0.32\linewidth}
    \includegraphics[width=\linewidth]{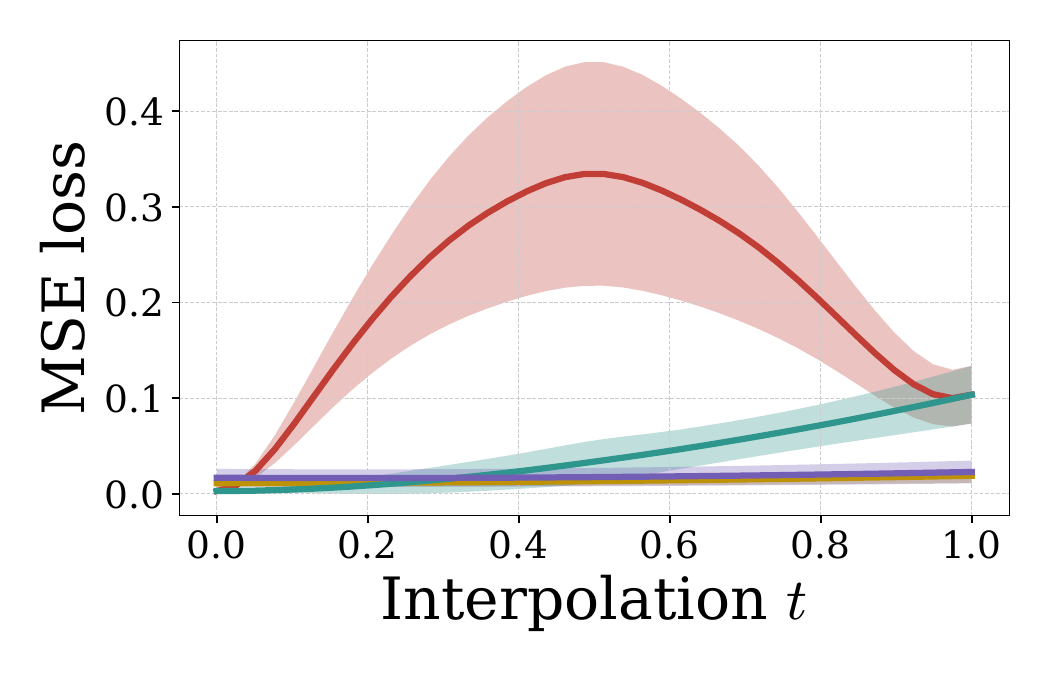}
    \caption{Permutations + Scaling}
  \end{subfigure}
  
  \caption{Interpolation experiments comparing different neural network alignment methods across various perturbation scenarios. }
  \label{fig:interpolation-inr}
  \vspace{-10pt}
\end{figure}

Alignment via Git Re-Basin shows a higher loss at the endpoints due to the perturbations introduced by the added noise. In contrast, both autoencoder variants are robust to this noise, with the ScaleGMN encoder performing slightly better, maintaining a lower loss profile. The ScaleGMN-based autoencoder performs marginally better than the Neural Graphs version, demonstrating the slight benefit of its scaling symmetry-aware architecture in this controlled setting. Similar qualitative behavior is observed across the three orbit-construction settings (permutation only, scaling only, and both), supporting the applicability of both our method and the Git Re-Basin algorithm.

% In addition, we interpolate directly in the latent space of the ScaleGMN autoencoder and obtain interpolation performance identical to its weight-space interpolation. Results are provided in Appendix (Section~\ref{appendix:interpolation-latent}).
Furthermore, when we interpolate directly in the ScaleGMN latent space for identical INR pairs, we observe identical performance to weight-space interpolation, confirming that our autoencoder learns robust representations (detailed results in Appendix Section~\ref{appendix:interpolation-latent}).

\begin{figure}[h!]
          \centering
  % --- Enhanced Legend with border ---
  \legendbox{%
    \legenditem{naivec}{Naive}\hspace{1.5em}%
    \legenditem{rebasin}{Linear Assignment}\hspace{1.5em}%
    \legenditem{ngraphs}{Neural Graphs}\hspace{1.5em}%
    \legenditem{scalegmn}{ScaleGMN}%
  }
  \vspace{4pt}
  
          \begin{subfigure}[b]{0.4\linewidth}
            \includegraphics[width=\linewidth]{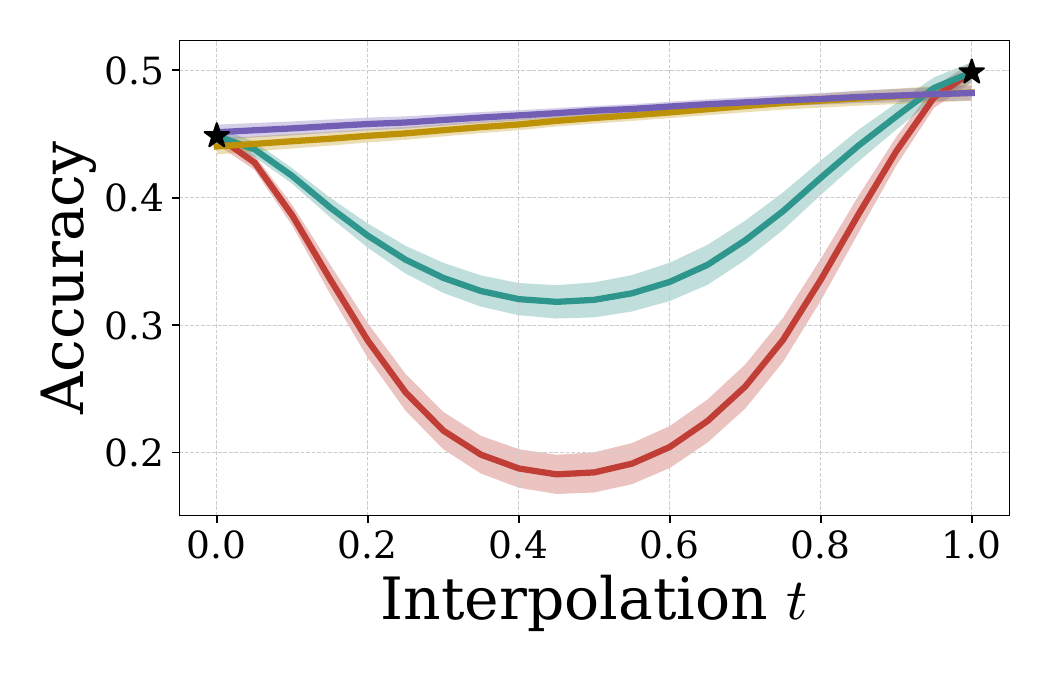}
            \caption{Accuracy}
          \end{subfigure}
          % \hspace{5pt}
          \begin{subfigure}[b]{0.4\linewidth}
            \includegraphics[width=\linewidth]{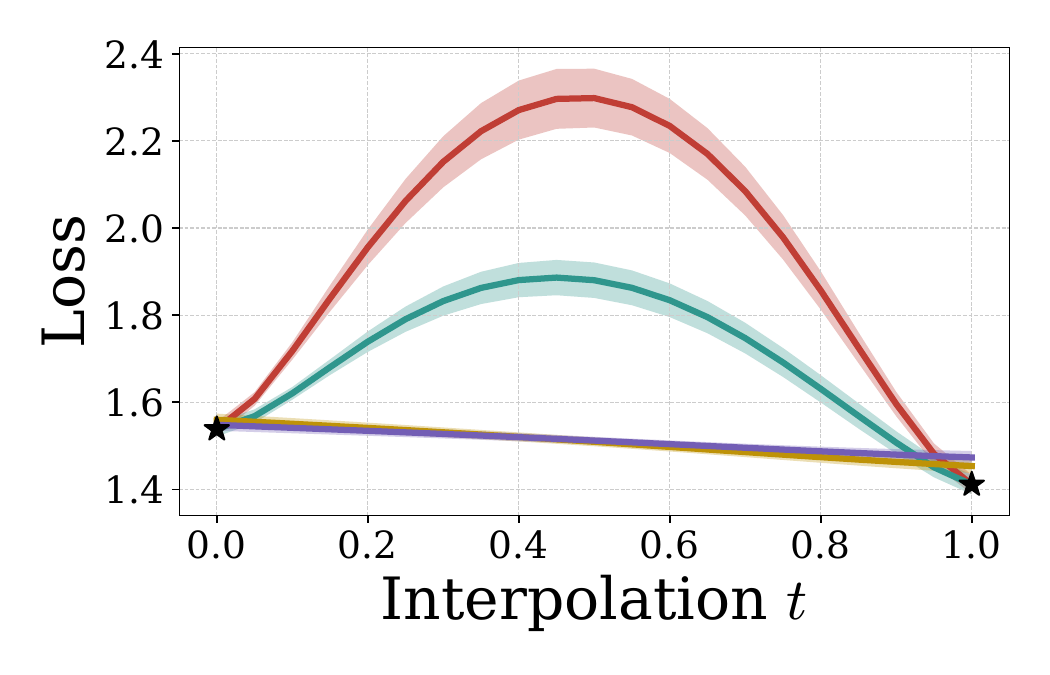}
            \caption{Cross Entropy Loss}
          \end{subfigure}
            \caption{Average interpolation curves (and standard deviation) over 20 pairs of distinct CNN models with ReLU activation.} 
          \label{fig:interpolation-cnn-relu-multiple-pairs}
        \end{figure}

\begin{figure}[h!]
          \centering
          \legendbox{%
    \legenditem{naivec}{Naive}\hspace{1.5em}%
    \legenditem{rebasin}{Linear Assignment}\hspace{1.5em}%
    \legenditem{ngraphs}{Neural Graphs}\hspace{1.5em}%
    \legenditem{scalegmn}{ScaleGMN}%
  }
  \vspace{4pt}
  
          \begin{subfigure}[b]{0.4\linewidth}
            \includegraphics[width=\linewidth]{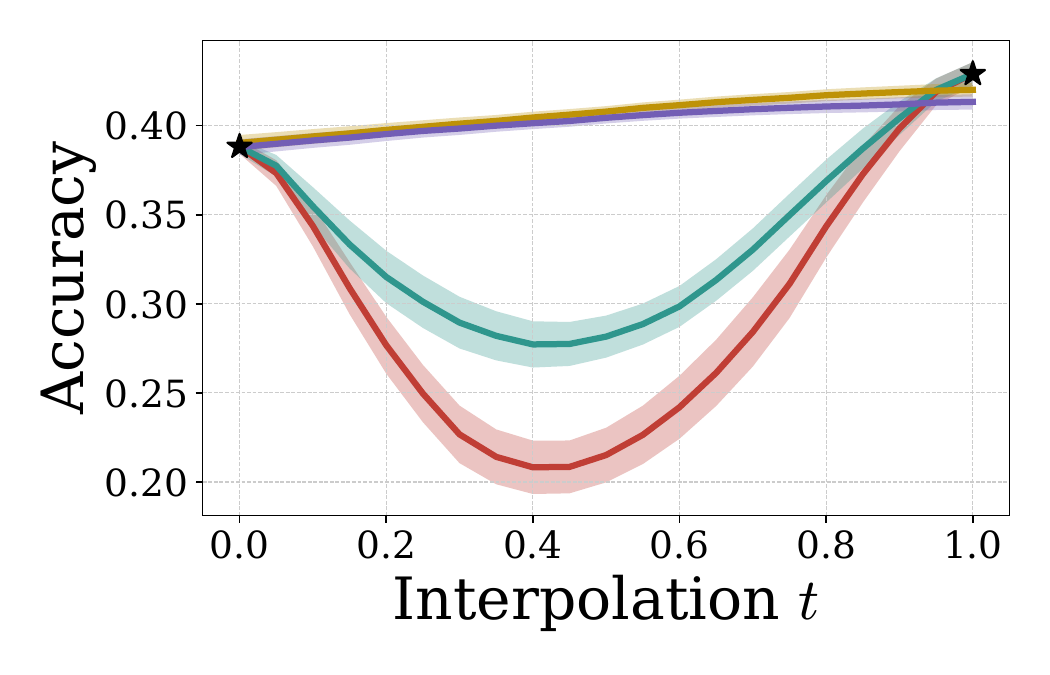}
            \caption{Accuracy}
          \end{subfigure}
          % \hspace{5pt}
          \begin{subfigure}[b]{0.4\linewidth}
            \includegraphics[width=\linewidth]{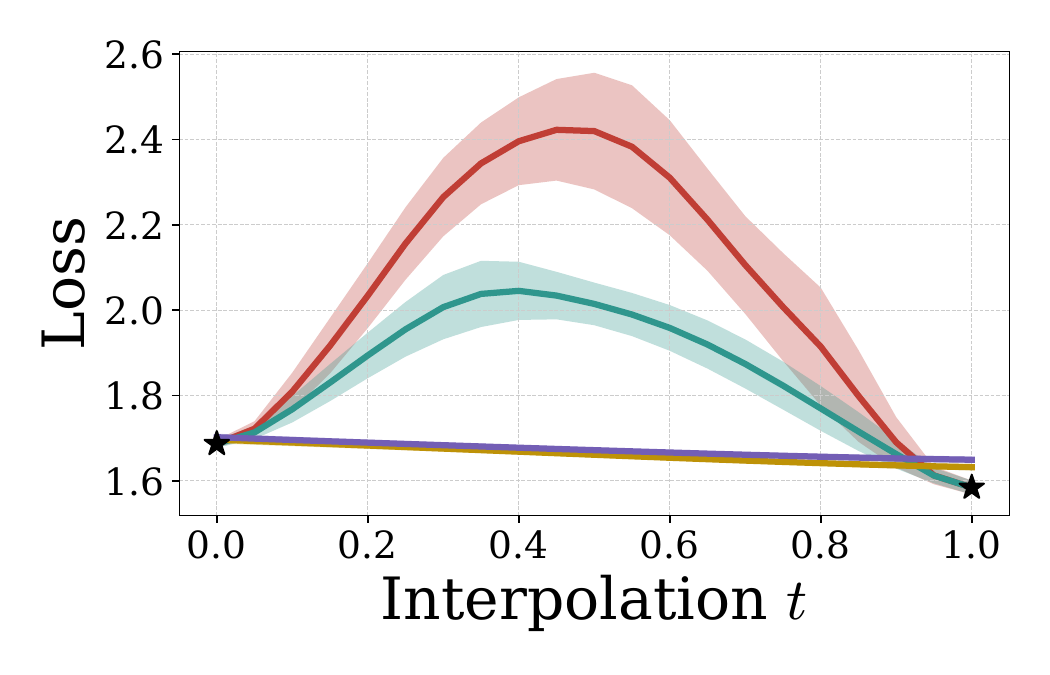}
            \caption{Cross Entropy Loss}
          \end{subfigure}
            \caption{Average interpolation curves (and standard deviation) over 20 pairs of distinct CNN models with Tanh activation.} 
          \label{fig:interpolation-cnn-tanh-multiple-pairs}
        \end{figure}

\paragraph{CNN Interpolation.}
The results of interpolating functionally distinct CNNs are presented in Figures~\ref{fig:interpolation-cnn-relu-multiple-pairs} and \ref{fig:interpolation-cnn-tanh-multiple-pairs}, which show average interpolation curves computed over 20 pairs of distinct CNN models. Additional results for individual CNN pairs are provided in Appendix (Section~\ref{appendix:cnn-interpolation-additional}). While Git Re-Basin substantially mitigates the catastrophic failure observed with naive interpolation, it still faces a significant performance barrier, with accuracy dropping sharply at the midpoint. This aligns with the findings of \citet{Ainsworth_2023_GitReBasin}, which indicate that permutation matching alone is often insufficient for connecting networks of limited width, such as those in the SmallCNN Zoo dataset. Furthermore, \citet{navon2023equivariant} report that the weight-matching strategy employed by Git Re-Basin is not optimal on CNN datasets. %To be clear, Git Re-Basin is highly effective in its designed domain; when aligning a network with a channel-permuted version of itself, it recovers the exact permutation and produces a perfectly flat interpolation line (see Appendix ).

In contrast, our autoencoder's reconstructed space enables a nearly monotonic, straight-line interpolation path between lower- and higher-performing models, revealing a comprehensive geometric representation that captures permutations and scaling symmetries. Furthermore, the comparable performance of the permutation-invariant Neural Graphs variant and the scale-aware ScaleGMN variant suggests that scaling symmetries hold less importance for model merging within the context of the datasets used.

\section{Discussion}
\subsection{Importance of scaling symmetries}
\label{sec:dis_imp_scale}

\begin{wrapfigure}{r}{0.4\textwidth} % 'r' = right, 'l' = left
  \centering
  \vspace{-30pt}
  \begin{subfigure}[b]{\linewidth}
    \includegraphics[width=\linewidth]{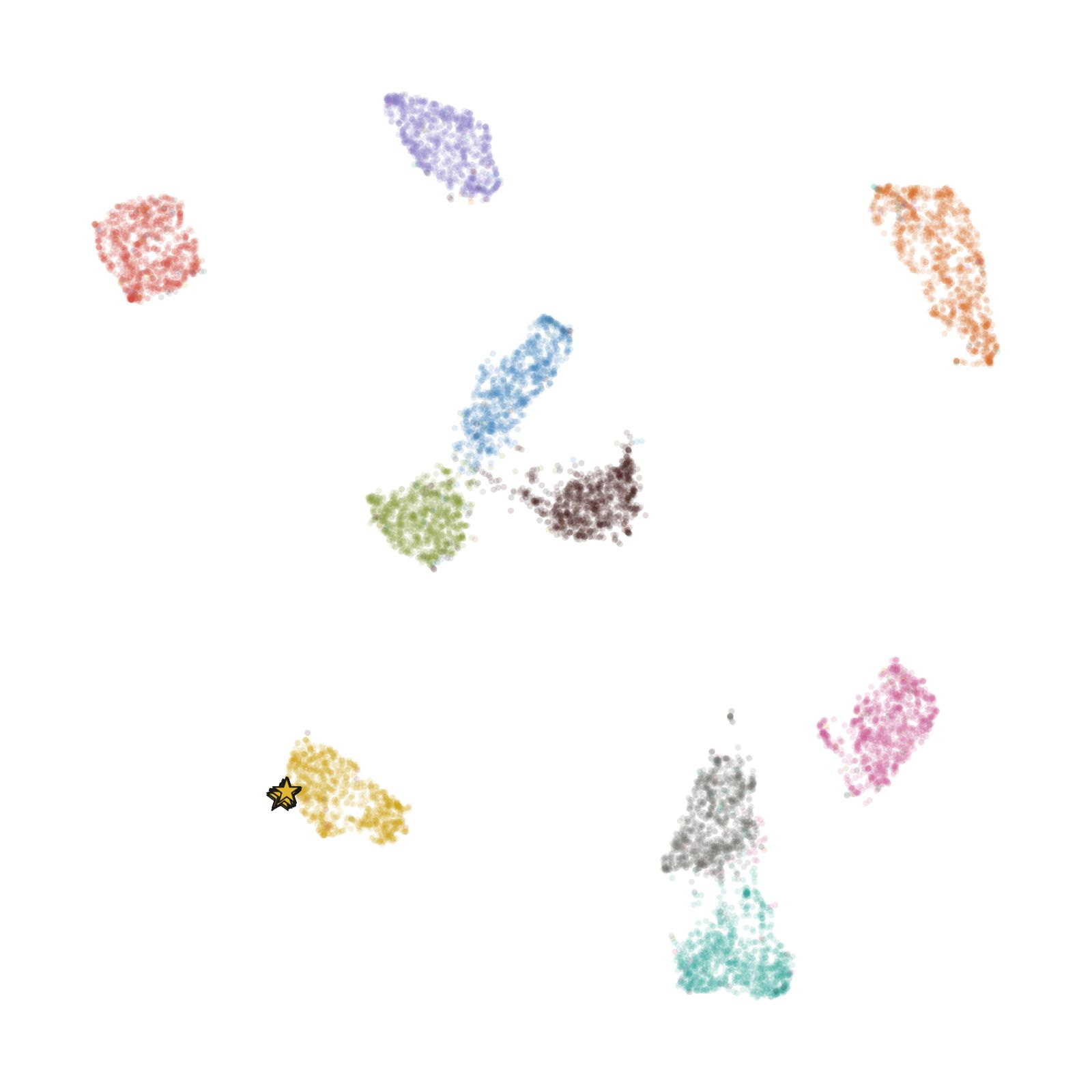}
    \caption{Scale‐GMN}
  \end{subfigure}%
  
  \begin{subfigure}[b]{\linewidth}
    \includegraphics[width=\linewidth]{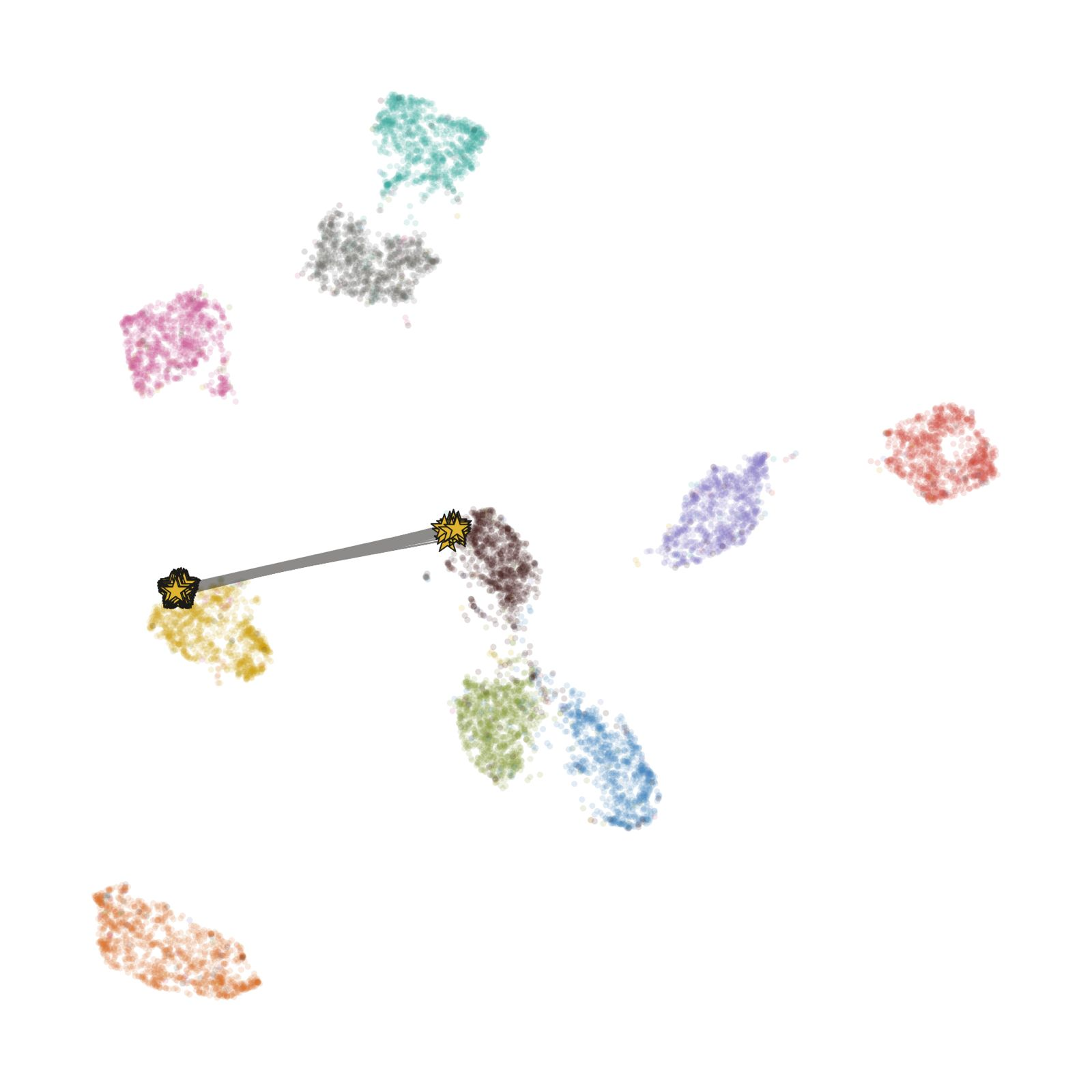}
    \caption{Neural Graphs}
  \end{subfigure}
  \caption{
    Latent representations learned by both encoder variants.
    The stars represent weight-transformed, functionally equivalent versions of the same INR.
  }
  \label{fig:umap}
  \vspace{-30pt}
\end{wrapfigure}

We begin by visualizing the latent representations learned by both the ScaleGMN and Neural Graphs encoders for the INR experiments (see Figure \ref{fig:umap}). Using UMAP \cite{McInnes_2020_UMAP} to reduce the dimensionality of latent representations across the entire INR dataset, we observe that data points naturally cluster by class (i.e., the digit they represent), despite the autoencoder being trained solely for reconstruction with functional equivalence rather than classification.

While the interpolation experiments presented in Section~\ref{section:results} demonstrate that model merging can be achieved in expectation without explicitly leveraging scaling symmetries, we further investigate the importance of addressing these symmetries within our autoencoder framework.

To assess the role of scaling symmetries, we plot the latent representation of a reference data point alongside a subset of its orbit generated with respect to the scaling group. As shown in Figure \ref{fig:umap}, the ScaleGMN encoder successfully collapses all orbit members to a single point in the latent space, whereas the Neural Graphs encoder fails to achieve this collapse. This difference has important implications: since the ScaleGMN latent space aligns all scale-equivalent networks to the same representation, their decoded versions naturally correspond to the same canonical representative in parameter space. In contrast, the Neural Graphs encoder's inability to collapse scale-equivalent networks in the latent space prevents it from producing canonicalized representations, making it impossible to recover scale-equivalent canonical networks from its outputs.

% \begin{figure}[ht!]
%   \centering
%   \begin{subfigure}[b]{0.45\linewidth}
%     \includegraphics[width=\linewidth]{media/importance_scaling/umap_scalegmn.png}
%     \caption{Scale‐GMN}
%   \end{subfigure}
%   \hspace{10pt}
%   \begin{subfigure}[b]{0.45\linewidth}
%     \includegraphics[width=\linewidth]{media/importance_scaling/umap_neural_graphs.png}
%     \caption{Neural Graphs}
%   \end{subfigure}\hfill
%   \caption{
%     Latent representations learned by both encoders variants.
%     The stars represent weight-transformed, thus still functionally equivalent, versions of the same INR.
% }
% \vspace{-10pt}
%   \label{fig:umap}
% \end{figure}

Therefore, while scale invariance may be less critical for model merging tasks specifically, it becomes essential for other latent space applications. Following approaches in \cite{DeLuigi_2023,Zhou_2023_NeuralFunctionalTransformers}, one could utilize these latent codes as network proxies for downstream tasks such as performance prediction, network classification, and INR-based generative modeling. For all such applications, having a latent representation where each point represents an entire family of functionally equivalent networks—as achieved by our ScaleGMN encoder—would be highly beneficial.

\subsection{Algorithmic complexity}
\label{sec:comlex}

The inference time complexity of the autoencoder scales linearly with the total number of parameters, $P$, due to its main components: the ScaleGMN encoder and MLP decoder. All model operations—initialization, message computation, aggregation, node and edge updates, and readout—scale with $\mathcal{O}(P)$. The encoder is dominated by message passing, with complexity $\mathcal{O}(L_{\mathrm{gnn}} P)$, where $L_{\mathrm{gnn}}$ is the number of GNN layers. Both graph initialization and readout, as well as the MLP decoder, scale linearly with $P$.
Linear assignment using the Hungarian algorithm has complexity $\mathcal{O}(n^3)$ per iteration for an $n \times n$ cost matrix. In the worst case, $n \approx \sqrt{P}$, giving a cost of $\mathcal{O}(P^{3/2})$ per iteration. With $T$ iterations, this becomes $\mathcal{O}(T \cdot P^{3/2})$, though in architectures with deep networks and fixed-size weight matrices (e.g., CNNs with small kernels), the cost is reduced to $\mathcal{O}(T \cdot P)$.
Comparing both methods, Git Re-Basin is an iterative method that requires multiple assignments until convergence and constitutes an approximation \cite{ruan2025task} to the original assignment problem  (see more in the Appendix, Section~\ref{appendix:rebasin-generalization}), whereas ScaleGMN and Neural Graph autoencoders require model training, which implies sufficient training iterations and data. In turn, they present a complexity of $\mathcal{O}(P)$ during inference, while Git Re-Basin presents $\mathcal{O}(T \cdot P)$ and $\mathcal{O}(T \cdot P^{3/2})$ in its best and worst cases, respectively.

\vspace{-5pt}
\section{Conclusion}
\vspace{-3pt}
% \paragraph{Summary.}
% We investigate the challenging problem of model merging, which is often hindered by parameter-space symmetries. The key to our approach is a symmetry-aware autoencoder framework that leverages ScaleGMNs to respect the natural permutation and scaling symmetries of the problem. Instead of explicitly solving a combinatorial assignment problem, our method learns to map functionally equivalent networks to a single canonical representation in weight space. At inference time, our method canonicalizes unseen networks in a single forward pass, bypassing iterative optimization. Our experiments demonstrate superior linear mode connectivity compared to both naive interpolation and explicit alignment methods. By using a Neural Graphs encoder as a comparative backbone, we also empirically investigate the importance of accounting for scaling symmetries on INRs and small CNNs. Furthermore, we show that the learned latent space is well-structured, suggesting its potential for downstream tasks beyond model merging.

% \paragraph{Limitations and future work.}
% One limitation of our approach is the need to train an autoencoder on a representative dataset of networks. Another limitation is that our empirical validation is confined to smaller-scale MLPs and CNNs. A key direction for future work is to investigate how this canonicalization pipeline scales to much larger and more complex architectures, such as Transformers and Large Language Models (LLMs). 

We address the challenging problem of model merging, which is often hindered by parameter-space symmetries. Our approach introduces a symmetry-aware autoencoder framework that leverages ScaleGMNs to respect the natural permutation and scaling symmetries inherent in neural networks. Rather than explicitly solving a combinatorial assignment problem, the method learns to map functionally equivalent networks to a single canonical representation in weight space.

At inference time, our framework canonicalizes unseen networks in a single forward pass, eliminating the need for iterative optimization. Experiments demonstrate superior linear mode connectivity compared to both naive interpolation and explicit alignment methods. Using a Neural Graphs encoder as a comparative backbone, we further highlight the importance of incorporating scaling symmetries alongside permutation symmetries on INRs and small CNNs. In addition, the learned latent space exhibits well-structured properties, suggesting potential for downstream applications beyond model merging.

This study primarily evaluates the framework on smaller-scale MLPs and CNNs, providing a controlled setting in which to establish its effectiveness. A limitation of the approach is the requirement to train the autoencoder on a representative dataset of networks---an upfront cost that enables efficient canonicalization at inference time. Together, these aspects form a strong foundation for demonstrating feasibility, while leaving open opportunities to extend the framework to larger and more complex architectures, such as Transformers and Large Language Models \cite{tran2024equivariant}. We view this as an exciting direction for future research, with the potential to enable practical, symmetry-aware canonicalization at scale.

\section*{Acknowledgements}
We wish to thank Erik J. Bekkers and Efstratios Gavves, as well as the University of Amsterdam, for providing access to the computational resources for this project.
Alejandro García Castellanos is funded by the Hybrid Intelligence Center, a 10-year programme funded through the research programme Gravitation which is (partly) financed by the Dutch Research Council (NWO).

\newpage
\bibliographystyle{plainnat}
\bibliography{refs}

\appendix

\newpage

\section{Generalized Git Re-Basin algorithm}
\label{appendix:rebasin-generalization}

As the authors in \cite{Ainsworth_2023_GitReBasin} state, the sum of a bilinear assignment problem is an NP-hard problem.
Similarly, we approximate the linear assignment problem (LAP) using their same formulation, where we also account for weight matching and a transformation consisting of permutation and scale $\boldsymbol{T} = \boldsymbol{P} \boldsymbol{Q}$.
Equation~\eqref{LAP_1} constitutes the starting objective for approximating the LAP, where $\langle \cdot \rangle_F$ denotes the Frobenius inner product.
Equation~\eqref{LAP_2} then follows from the definition of the Frobenius inner product and properties of the trace.
\resizebox{\linewidth}{!}{%
\begin{minipage}{\linewidth}
\begin{align}
& \arg \max_{\boldsymbol{T}_{\ell}} \left( \langle \boldsymbol{W}_{\ell}^{(\mathrm{A})}, \boldsymbol{T}_{\ell} \boldsymbol{W}_{\ell}^{(\mathrm{B})} \boldsymbol{T}_{\ell-1}^{\top} \rangle_F
+ \langle \boldsymbol{W}_{\ell+1}^{(\mathrm{A})}, \boldsymbol{T}_{\ell+1} \boldsymbol{W}_{\ell+1}^{(\mathrm{B})} \boldsymbol{T}_{\ell}^{\top} \rangle_F
+ \langle \boldsymbol{b}_{\ell}^{(\mathrm{A})}, \boldsymbol{T}_{\ell} \boldsymbol{b}_{\ell}^{(\mathrm{B})} \rangle_F \right) \label{LAP_1} \\
&= \arg \max_{\boldsymbol{T}_{\ell}} \left( \mathrm{Tr}\left( \boldsymbol{T}_{\ell} \boldsymbol{W}_{\ell}^{(\mathrm{B})} \boldsymbol{T}_{\ell-1}^{\top} (\boldsymbol{W}_{\ell}^{(\mathrm{A})})^{\top} \right)
+ \mathrm{Tr}\left( \boldsymbol{T}_{\ell} (\boldsymbol{W}_{\ell+1}^{(\mathrm{B})})^{\top} \boldsymbol{T}_{\ell+1}^{\top} \boldsymbol{W}_{\ell+1}^{(\mathrm{A})} \right)
+ \mathrm{Tr}\left( \boldsymbol{T}_{\ell} \boldsymbol{b}_{\ell}^{(\mathrm{B})} (\boldsymbol{b}_{\ell}^{(\mathrm{A})})^{\top} \right) \right) \\
&= \arg \max_{\boldsymbol{T}_{\ell}} \mathrm{Tr}\left( \boldsymbol{T}_{\ell} \left( \boldsymbol{W}_{\ell}^{(\mathrm{B})} \boldsymbol{T}_{\ell-1}^{\top} (\boldsymbol{W}_{\ell}^{(\mathrm{A})})^{\top}
+ (\boldsymbol{W}_{\ell+1}^{(\mathrm{B})})^{\top} \boldsymbol{T}_{\ell+1}^{\top} \boldsymbol{W}_{\ell+1}^{(\mathrm{A})}
+ \boldsymbol{b}_{\ell}^{(\mathrm{B})} (\boldsymbol{b}_{\ell}^{(\mathrm{A})})^{\top} \right) \right) \\
&= \arg \max_{\boldsymbol{T}_{\ell}} \langle \boldsymbol{T}_{\ell}, \underbrace{\left( \boldsymbol{W}_{\ell}^{(\mathrm{A})} \boldsymbol{T}_{\ell-1} (\boldsymbol{W}_{\ell}^{(\mathrm{B})})^{\top}
+ (\boldsymbol{W}_{\ell+1}^{(\mathrm{A})})^{\top} \boldsymbol{T}_{\ell+1} \boldsymbol{W}_{\ell+1}^{(\mathrm{B})}
+ \boldsymbol{b}_{\ell}^{(\mathrm{A})} (\boldsymbol{b}_{\ell}^{(\mathrm{B})})^{\top} \right)}_{:= \; \boldsymbol{C}_{\ell}} \rangle_F \label{LAP_2}
\end{align}
Defining $\boldsymbol{P}_{\ell} = (p^{(\ell)}_{ij})_{1 \le i,j \le N}$ and $\boldsymbol{Q}_{\ell} = \mathrm{diag}(q^{(\ell)}_1, \hdots, q^{(\ell)}_N)$ and rewriting the objective function by explicitly writing the transformation as $\boldsymbol{T} = \boldsymbol{P} \boldsymbol{Q}$ we obtain
\begin{align} \label{LAP_3}
\left\langle \boldsymbol{P}_{\ell} \boldsymbol{Q}_{\ell}, \boldsymbol{C}_{\ell} \right\rangle_F
= \sum_{i=1}^{N} \sum_{k=1}^{N} (p^{(\ell)}_{ik} q^{(\ell)}_k) (\boldsymbol{C}_{\ell})_{ik}
= \sum_{i=1}^{N} q^{(\ell)}_{\pi(i)} (\boldsymbol{C}_{\ell})_{i, \pi(i)}.
\end{align}
\end{minipage}
}
We now demonstrate that when the network's activation functions are either sine or $\tanh$, this optimization problem admits an efficient solution within a single algorithm iteration. The key insight is to decompose the problem into two sequential steps: first solve an approximate Linear Assignment Problem (LAP) to obtain $\boldsymbol{P}_{\ell}$, then determine $\boldsymbol{Q}_{\ell}$ as an explicit function of both $\boldsymbol{C}_{\ell}$ and the previously computed $\boldsymbol{P}_{\ell}$.

\begin{proposition} \label{proposition_1}
Suppose the activation functions of the network are either $\sin$ or $\tanh$. Then, for a fixed cost matrix $\boldsymbol{C}_{\ell}$, the optimal transformation $\boldsymbol{T}_{\ell} = \boldsymbol{P}^*_{\ell} \boldsymbol{Q}^*_{\ell}$ that maximizes the objective in Equation~\eqref{LAP_3} is given by:
\begin{itemize}
    \item[$\boldsymbol{\cdot}$] $\boldsymbol{Q}^*_{\ell} = \mathrm{diag}({q^*}^{(\ell)}_1, \ldots, {q^*}^{(\ell)}_N)$, where ${q^*}^{(\ell)}_{i} = \mathrm{sign}((\boldsymbol{C}_{\ell})_{i, \pi^*(i)})$, and
    \item[$\boldsymbol{\cdot}$] $\boldsymbol{P}^*_{\ell} = \underset{\boldsymbol{P}_{\ell}}{\arg\max} \sum_{i=1}^{N} |(\boldsymbol{C}_{\ell})_{i, \pi(i)}|$.
\end{itemize}
\end{proposition}

\begin{proof}
For an arbitrary transformation $\boldsymbol{P}_{\ell} \boldsymbol{Q}_{\ell}$,
\begin{align}
\left\langle \boldsymbol{P}^*_{\ell} \boldsymbol{Q}^*_{\ell}, \boldsymbol{C}_{\ell} \right\rangle_F
&= \sum_{i=1}^{N} \mathrm{sign}((\boldsymbol{C}_{\ell})_{i, \pi^*(i)}) (\boldsymbol{C}_{\ell})_{i, \pi^*(i)} \\
&= \sum_{i=1}^{N} |(\boldsymbol{C}_{\ell})_{i, \pi^*(i)}| \\
&\geq \sum_{i=1}^{N} |(\boldsymbol{C}_{\ell})_{i, \pi(i)}|
    \qquad\qquad\qquad \left(\text{permutation assignment $\pi^*(\cdot)$ is optimal}\right)\\
&\geq \sum_{\i=1}^{N} q^{(\ell)}_{\pi(i)} (\boldsymbol{C}_{\ell})_{i, \pi(i)}
    \qquad\qquad\qquad\qquad\quad\;\;  (q^{(\ell)}_{\pi(i)} \in \{+1, -1\} \text{ for all } \ell, i)\\
&= \left\langle \boldsymbol{P}_{\ell} \boldsymbol{Q}_{\ell}, \boldsymbol{C}_{\ell} \right\rangle_F.
\end{align}
\end{proof}

Essentially, Proposition \ref{proposition_1} shows that it suffices to perform weight matching considering only permutations on the absolute value of the cost matrix and later defining $\boldsymbol{Q}^*_{\ell}$ such that all terms in the sum of the inner product $\left\langle \boldsymbol{P}^*_{\ell} \boldsymbol{Q}^*_{\ell}, \boldsymbol{C}_{\ell} \right\rangle_F$ are positive, thus contribute to maximizing the objective.
This allows for an efficient weight matching approximation as opposed to considering the cartesian product of the groups of permutations and scalings.
It is important to note that $\boldsymbol{P}^*_{\ell} \boldsymbol{Q}^*_{\ell}$ are optimal for a fixed cost matrix $\boldsymbol{C}_\ell$, meaning that within a given iteration of the algorithm we can obtain the optimal transformation, but this may not be optimal with respect to the true objective.
Since $\boldsymbol{C}_\ell$ is a function of $\boldsymbol{P}_{\ell} \boldsymbol{Q}_{\ell}$, the obtained solution is an approximation, which we empirically show to have considerable standard deviation but behaves adequately in expectation (monotonically increasing and slightly convex LAP interpolation, Figure~\ref{fig:interpolation-inr}).

An easy counterexample to see how this method does not converge to the global optimum is to consider two INRs, where one was generated by just permuting but not flipping signs, this is, $\boldsymbol{Q}_{\ell} = \boldsymbol{I}$.
Now, $\boldsymbol{Q}^*_{\ell}$ will not necessarily be $\boldsymbol{I}$, since $\boldsymbol{C}_{\ell}$ can have negative entries, meaning the method will unnecessarily flip the signs of the INR and thus not match both into the same basin.
In this sense, there is no preference for setting $\boldsymbol{Q}_{\ell} = \boldsymbol{I}$, nor is the case for $\boldsymbol{P}_{\ell} = \boldsymbol{I}$.

Performing this process of optimization by fixing the cost matrices, $\boldsymbol{C}_{\ell}$, intra-iteration and by updating the transformations of one layer at a time and at random, we arrive at Algorithm \ref{alg:perm‐scale‐coord‐descent}.

\begin{algorithm}[ht!]
\caption{\textsc{Permutation+ScalingCoordinateDescent}}
\label{alg:perm‐scale‐coord‐descent}
\begin{algorithmic}[1]
\REQUIRE 
  Model weights 
  \(\boldsymbol{\Theta}_A = \{\boldsymbol{W}^{(A)}_1,\,\dots,\,\boldsymbol{W}^{(A)}_L\}\) 
  and 
  \(\boldsymbol{\Theta}_B = \{\boldsymbol{W}^{(B)}_1,\,\dots,\,\boldsymbol{W}^{(B)}_L\}.\)
\ENSURE 
  A set of layer‐wise transformations 
  \(\displaystyle \{\boldsymbol{T}_1,\,\dots,\,\boldsymbol{T}_{L-1}\}\), 
  where each \(\boldsymbol{T}_\ell = \boldsymbol{Q}_\ell\,\boldsymbol{P}_\ell\) is a composition of a permutation \(\boldsymbol{P}_\ell\) and a diagonal scaling \(\boldsymbol{Q}_\ell\), 
  chosen so as to approximately maximize 
  \(\mathrm{vec}(\boldsymbol{\Theta}_A)\,\cdot\,\mathrm{vec}\bigl(T(\boldsymbol{\Theta}_B)\bigr).\)

\STATE Initialize \(\boldsymbol{T}_\ell \leftarrow \boldsymbol{I}\) \quad for all \(\ell = 1,\dots,L-1.\)
\REPEAT
  \FOR{\(\ell \in \text{RandomPermutation}(1,\dots,L-1)\)}
    \STATE 
    \[
      \boldsymbol{T}_\ell \;\leftarrow\; 
      \textsc{SolveLAP}\Bigl(
        \boldsymbol{W}^{(A)}_\ell \,\boldsymbol{T}_{\ell-1}\,\bigl(\boldsymbol{W}^{(B)}_\ell\bigr)^\top \;+\;
        \bigl(\boldsymbol{W}^{(A)}_{\ell+1}\bigr)^\top \,\boldsymbol{T}_{\ell+1}\,\boldsymbol{W}^{(B)}_{\ell+1} + \boldsymbol{b}_\ell^{(A}(\boldsymbol{b}_\ell^{(B)})^T 
      \Bigr)
    \]

  \ENDFOR
\UNTIL{convergence}
\end{algorithmic}
\end{algorithm}

A critical analysis of parameter matching algorithms for model merging reveals fundamental trade-offs between the generality of the transformation and the optimality of the optimization strategy.
The authors in \cite{Zhang_2025_Transformers} propose a matching methodology that is constrained by its sequential optimization strategy, where rotation and rescaling symmetries are addressed in two separate, consecutive steps. This decoupled approach is fundamentally a greedy, path-dependent algorithm; by first committing to an optimal rotation matrix $R$, it constrains the subsequent search for a rescaling factor $a$. Consequently, it does not guarantee convergence to a global optimum, as a different initial rotation might have enabled a superior overall alignment. The authors of \cite{Zhang_2025_Transformers} acknowledge this deficiency, framing their method as a practical approximation that trades theoretical optimality for computational tractability.

In contrast, our proposed approach addresses the related problem of matching models under transformations composed of permutations and sign-flips. A crucial distinction lies in the optimization strategy: whereas the former method is sequential, our approach intertwines the optimization of permutations and scaling within each iteration. For a fixed cost matrix at a given step, it jointly computes the optimal permutation and sign-flips. However, this method's form of scaling is restricted to discrete values of $+1$ and $-1$, a notable limitation compared to the arbitrary, continuous scaling admitted by the former. The comparison highlights a central challenge in the field: one approach handles a more general scaling problem with a suboptimal sequential strategy, while the other employs a more principled intertwined strategy for a more restricted problem, yet both ultimately yield approximations.
This underscores the need for a more general and powerful approach to model merging that can accommodate arbitrary symmetries with formal guarantees of optimality, an effort the authors of \cite{Zhang_2025_Transformers} note is non-trivial for their joint optimization problem.

\section{Graph Metanetworks}\label{appendix:graph_metanetworks}

        %A \emph{Graph Metanetwork} \cite{Kofinas_2024_Graph} treats a FFNN’s topology as a graph $G=(V,E)$ where each node $v_{\ell,i}$ carries feature $f_v=b_{\ell,i}$ and each edge $(u\to v)$ carries $f_{u\to v}=W_{\ell}(i,j)$.  A $T$-step message passing is:
        Graph Metanetworks (GMNs) \cite{Kofinas_2024_Graph} process FFNNs  as graphs using conventional GNNs, leveraging permutation symmetries. The computational graph $G = (\mathcal{V}, \mathcal{E})$ uses $i \in \mathcal{V}$ for neurons and $(i, j) \in \mathcal{E}$ for edges from $j$ to $i$. Vertex features $\mathbf{x}_V \in \mathbb{R}^{|\mathcal{V}|\times d_v}$ (biases) and edge features $\mathbf{x}_E \in \mathbb{R}^{|\mathcal{E}|\times d_e}$ (weights) are inputs. A $T$-iteration GMN is defined as:
        \resizebox{\textwidth}{!}{%
\begin{minipage}{\textwidth}
        \begin{align*}
          &\mathbf{h}_V^0(i) ={}  \mathrm{INIT}_V (\mathbf{x}_V (i)), \quad\quad \mathbf{h}_E^0(i, j) = \mathrm{INIT}_E (\mathbf{x}_E (i, j)) && \text{(Init)} \\
          &\mathbf{m}_V^t(i) ={}  \bigoplus_{j \in N(i)} \mathrm{MSG}_V^t (\mathbf{h}_V^{t-1}(i), \mathbf{h}_V^{t-1}(j), \mathbf{h}_E^{t-1}(i, j)) && \text{(Msg)} \\
          &\mathbf{h}_V^t(i) ={}  \mathrm{UPD}_V^t (\mathbf{h}_V^{t-1}(i), \mathbf{m}_V^t(i)), \quad \quad\mathbf{h}_E^t(i, j) = \mathrm{UPD}_E^t (\mathbf{h}_V^{t-1}(i), \mathbf{h}_V^{t-1}(j), \mathbf{h}_E^{t-1}(i, j)) && \text{(Upd)} \\
          &\mathbf{h}_G ={}  \mathrm{READ} (\{\mathbf{h}_V^T (i)\}_{i \in V}) && \text{(Readout)}
        \end{align*}
\end{minipage}
}

        %\[
        %\begin{aligned}
        %  h_v^{(0)} 
        %    &= \mathrm{Init}_v(f_v), 
        %    &h_{u\to v}^{(0)}= \mathrm{Init}_{u\to v}(f_{u\to v})
        %    \\
        %  m_{u\to v}^{(t)} 
        %    &= \mathrm{MSG}\!\Bigl(h_u^{(t)}, f_{u\to v}\Bigr), 
        %    &h_v^{(t+1)} 
        %    = \mathrm{UPD}_v\!\Bigl(h_v^{(t)}, \sum_{u:\,u\to v} m_{u\to v}^{(t)}\Bigr), \,
        %    \\[0.3em]
        %  h_{u\to v}^{(t+1)} &=\mathrm{UPD}_{u\to v}\!\Bigl(h_u^{(t)}, h_v^{(t)},  h^{(t)}_{u\to v}\Bigr)
        %  &h_G 
        %    = \mathrm{READ}\Bigl(\{h_v^{(T)}\}_{v\in V}\Bigr).
        %\end{aligned}
        %\]

        where \textsc{READ} is optional and is usually permutation-invariant (e.g.\ DeepSets). 

     \paragraph{Scale-Equivariant Message Passing:}
        \label{section:message_passing}
        In order to respect scaling symmetries, Kalogeropoulos et al.~\cite{Kalogeropoulos_2024} replace $\mathrm{MSG}$ and $\mathrm{UPD}$ of the standard Graph Metanetwork framework with \emph{scale-equivariant} blocks.  Define:
        \begin{align}
          \mathrm{ScaleEq}(x_1,\dots,x_n)
          &=\bigl[\Gamma_1x_1,\dots,\Gamma_nx_n\bigr]
            \;\odot\;
            \rho\bigl(\mathrm{canon}(x_1),\dots,\mathrm{canon}(x_n)\bigr),
          \label{eq:scaleeq}\\
          \mathrm{ReScaleEq}(y,e)
          &=(\Gamma_y\,y)\;\odot\;(\Gamma_e\,e),
          \label{eq:rescale}
        \end{align}
        where the canonicalization function
        $\mathrm{canon}(x)$ is chosen based on the underlying symmetry. For \textbf{sign symmetry},  $\mathrm{canon}(x) = \mathrm{MLP}(x) + \mathrm{MLP}(-x)$ is used whereas for \textbf{positive scalings}, L2 normalization $\mathrm{canon}(x) = x / \|x\|$ is used. The $\mathrm{canon}(x)$ function removes scale, thus making $\rho$ invariant, and $\Gamma_i$ are learnable linear maps which are then combined with the scale invariant component to achieve scale equivariance.  
        Then
        \[
          \mathrm{MSG}_{\rm SE}(\mathbf x,\mathbf y,\mathbf e)
          = \mathrm{ScaleEq}\bigl[\mathbf x,\mathrm{ReScaleEq}(\mathbf y,\mathbf e)\bigr],
          \quad
          \mathrm{UPD}_{\rm SE}(\mathbf x,\mathbf m)
          = \mathrm{ScaleEq}\bigl[\mathbf x,\mathbf m\bigr].
        \]
        For any diagonal $D_\ell\succ0$,
        $
          \mathrm{MSG}_{\rm SE}(D_\ell x,\,D_{\ell-1}y,\,D_\ell\,D_{\ell-1}^{-1}e)
          = D_\ell\,\mathrm{MSG}_{\rm SE}(x,y,e),
        $
        and similarly for $\mathrm{UPD}_{\rm SE}$, thus satisfying equivariance for the scaling subgroup with respect to the central node $x$.  Combining these with the permutation-equivariant GMN yields a model equivariant to the full group of scalings and permutations.

\section{Experimental details}
\label{appendix:experimental-details}

\subsection{Datasets and splits}
\label{appendix:data-splits}

Our experiments utilize the following dataset partitions. For the Implicit Neural Representation (INR) models, the train/validation/test split was 55,000/5,000/10,000 samples. For the Convolutional Neural Network (CNN) models, the splits varied by activation function: 5,976/1,465/7,433 for the Tanh variant and 6,024/1,535/7,567 for the ReLU variant.

The CNN models' functional loss—defined as the KL divergence between the output distributions of the original and reconstructed networks—was computed using images from the CIFAR-10 dataset. To ensure computational efficiency during training, we used a fixed, random subset of 10,000 CIFAR-10 images. For evaluation tasks such as generating interpolation lines, a distinct, held-out set of 10,000 images was used.

\subsection{Hyperparameters}
\label{appendix:hypeparam-sweep}

To ensure optimal performance, we conducted a comprehensive hyperparameter sweep for our primary models. The search spaces for the key hyperparameters for our Implicit Neural Representation (INR) and Convolutional Neural Network (CNN) autoencoder architectures are detailed in Table~\ref{tab:inr-hyperparams} and Table~\ref{tab:cnn-hyperparams}, respectively. The final hyperparameter values chosen for our experiments, which yielded the best performance on a smaller portion of the validation set, are indicated in bold. Further details on the final graph neural network architectures and optimizer settings for both the INR and CNN experiments are provided in Table~\ref{tab:inr-final-params} and Table~\ref{tab:cnn-final-params}.

\begin{table}[ht!]
  \caption{Hyperparameters explored for INR experiments. The decoder hidden dimensions are computed from a base hidden size of 128.}
  \label{tab:inr-hyperparams}
  \medskip
  \centering
  \begin{tabular}{ll}
    \toprule
    \textbf{Hyperparameter} & \textbf{Search Space} \\
    \midrule
    Decoder hidden layers & [],\; [256],\; \textbf{[256, 512]},\; [256, 512, 1024] \\
    Learning rate & 0.01,\; \textbf{0.001},\; 0.0001 \\
    \bottomrule
  \end{tabular}
\end{table}

\begin{table}[ht!]
  \caption{Hyperparameters explored for CNN experiments. Decoder hidden dimensions are computed from the base hidden size (128 or 256).}
  \label{tab:cnn-hyperparams}
  \medskip
  \centering
  \begin{tabular}{ll}
    \toprule
    \textbf{Hyperparameter} & \textbf{Search Space} \\
    \midrule
    Temperature & \textbf{0.5},\; 1.0,\; 1.5 \\
    Learning rate & \textbf{0.001},\; 0.0005,\; \textbf{0.0001} \\
    Encoder Backbone hidden dimension & 128,\; \textbf{256} \\
    Decoder hidden layers (base=128) & [256, 512],\; [256, 512, 1024] \\
    Decoder hidden layers (base=256) & \textbf{[512, 1024]},\; [512, 1024, 2048] \\
    \bottomrule
  \end{tabular}
\end{table}

\begin{table}[ht!]
\caption{Final model and optimizer parameters for the INR experiments.}
\medskip
\label{tab:inr-final-params}
\centering
\begin{tabular}{lcc}
\toprule
\textbf{Hyperparameter} & \textbf{ScaleGMN} & \textbf{NeuralGraphs} \\
\midrule
Epochs & 300 & 300 \\
Optimizer & AdamW & AdamW \\
Learning Rate & $1 \times 10^{-3}$ & $1 \times 10^{-3}$ \\
Learning Rate Warmup Steps & 1K & 1K \\
Weight Decay & $1 \times 10^{-2}$ & $5 \times 10^{-4}$ \\
GNN Layers & 4 & 4 \\
Hidden Dimension & 128 & 128 \\
Readout Method & Full graph & Last-layer pooling \\
Decoder Hidden Layers & [256, 512] & [256, 512] \\
Decoder Activation & SiLU & SiLU \\
\bottomrule
\end{tabular}
\end{table}

\begin{table}[ht!]
\caption{Final model and optimizer parameters for the CNN experiments.}
\label{tab:cnn-final-params}
\medskip
\centering
\begin{tabular}{lccc}
\toprule
\textbf{Hyperparameter} & \textbf{ScaleGMN (ReLU)} & \textbf{ScaleGMN (Tanh)} & \textbf{NeuralGraphs} \\
\midrule
Epochs & 300 & 300 & 300 \\
Optimizer & AdamW & AdamW & AdamW \\
Learning Rate & $1 \times 10^{-3}$ & $1 \times 10^{-3}$ & $1 \times 10^{-4}$ \\
Learning Rate Warmup Steps & 1K & 1K & 1K \\
Weight Decay & $1 \times 10^{-2}$ & $1 \times 10^{-2}$ & $5 \times 10^{-4}$ \\
GNN Layers & 4 & 4 & 4 \\
Hidden Dimension & 256 & 256 & 256 \\
Readout Method & Full graph & Full graph & Last-layer pooling \\
Decoder Hidden Layers & [512, 1024] & [512, 1024] & [512, 1024] \\
Decoder Activation & SiLU & SiLU & SiLU \\
\bottomrule
\end{tabular}
\end{table}

%\section{Technical Appendices and Supplementary Material}
%Technical appendices with additional results, figures, graphs and proofs may be submitted with the paper submission before the full submission deadline (see above), or as a separate PDF in the ZIP file below before the supplementary material deadline. There is no page limit for the technical appendices.

\newpage
\section{Additional results}
This section presents additional results that support the claims made in this study. We also include a complete set of experiments to provide a comprehensive view of the proposed method.

\subsection{INR interpolation without noise}
\label{appendix:interpolation-without-noise}

\begin{figure}[h!]
    \centering
    \legendbox{%
    \legenditem{naivec}{Naive}\hspace{1.5em}%
    \legenditem{rebasin}{Linear Assignment}\hspace{1.5em}%
    \legenditem{ngraphs}{Neural Graphs}\hspace{1.5em}%
    \legenditem{scalegmn}{ScaleGMN}%
    }
    
    \vspace{4pt}

    % --- Your panels ---
    \begin{subfigure}[b]{0.32\linewidth}
    \includegraphics[width=\linewidth]{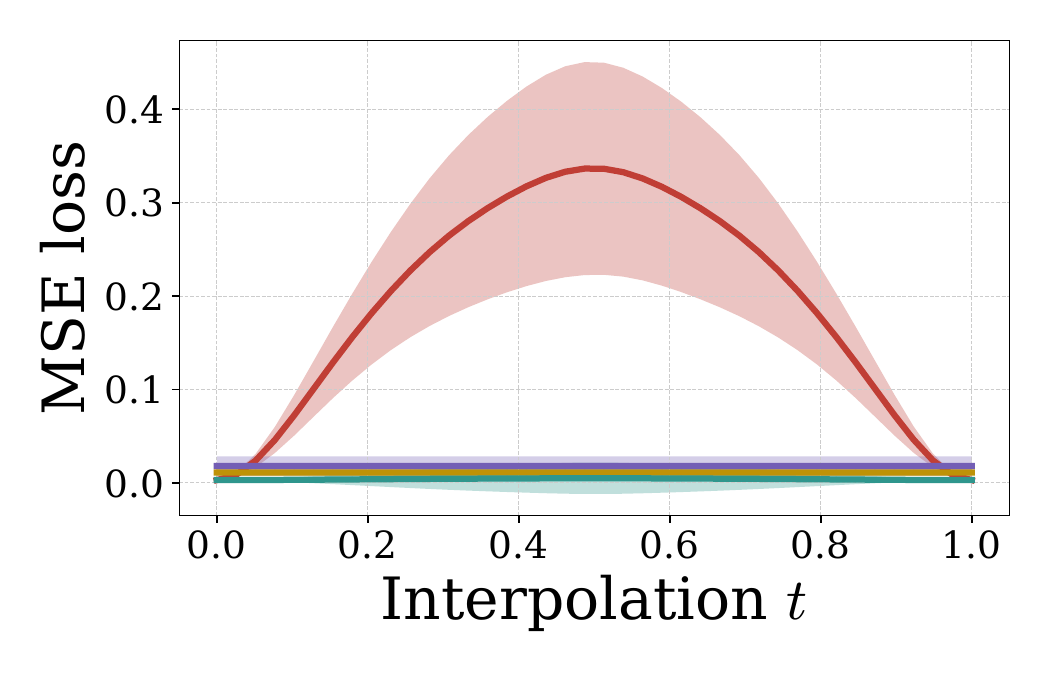}
    \caption{Permutations}
    \label{fig:interpolation-inr-without-noise:a}
    \end{subfigure}\hfill
    \begin{subfigure}[b]{0.32\linewidth}
    \includegraphics[width=\linewidth]{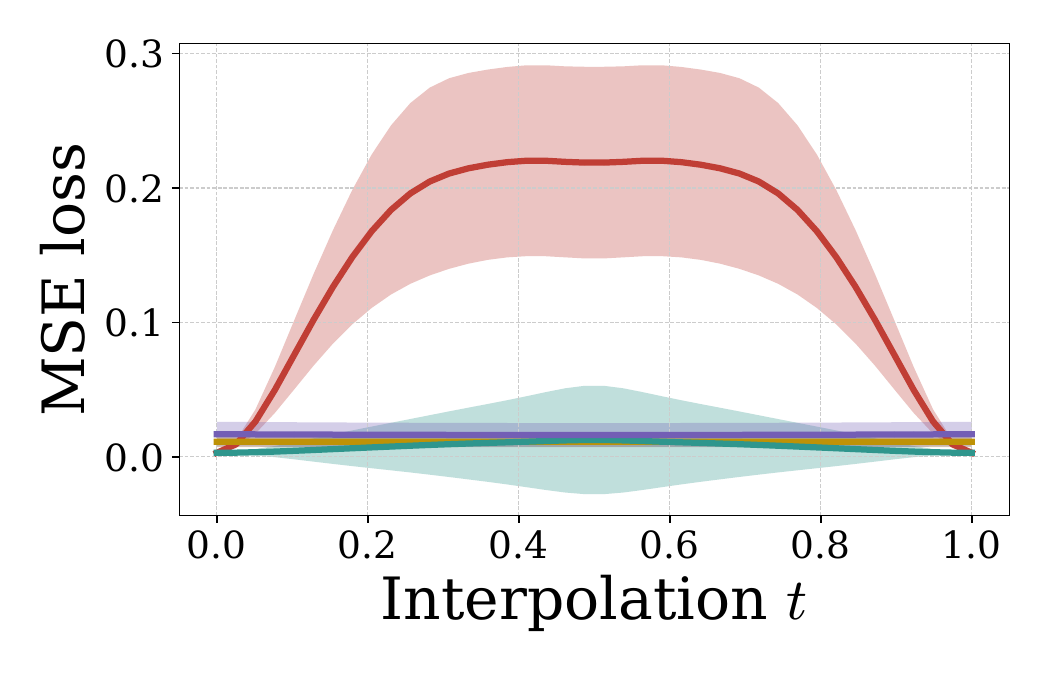}
    \caption{Scaling}
    \label{fig:interpolation-inr-without-noise:b}
    \end{subfigure}\hfill
    \begin{subfigure}[b]{0.32\linewidth}
    \includegraphics[width=\linewidth]{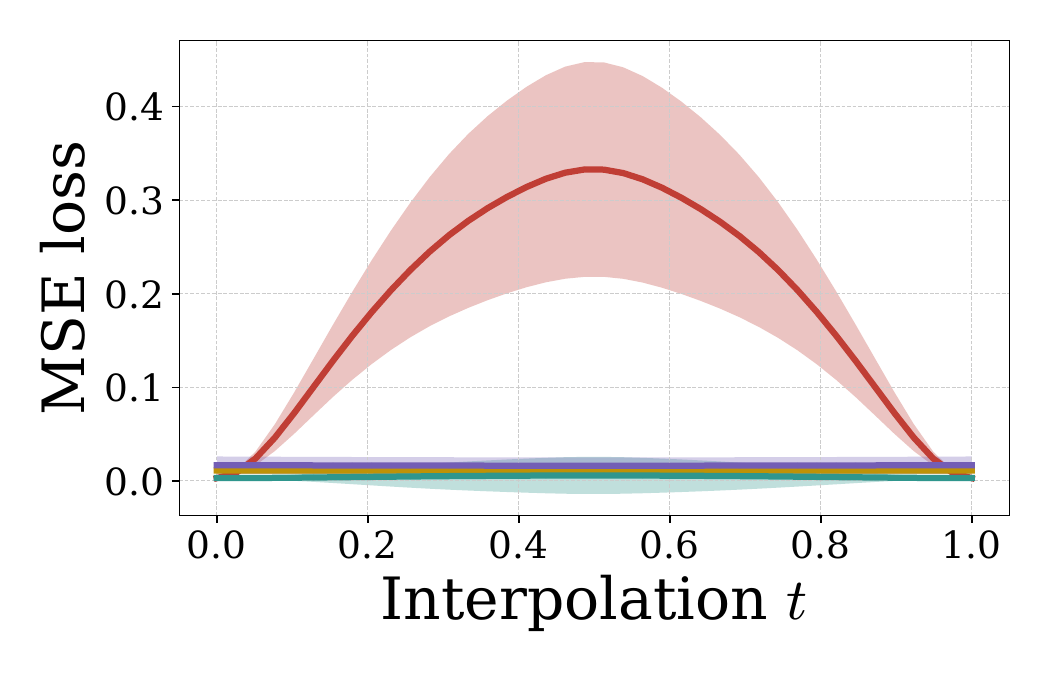}
    \caption{Permutations + Scaling}
    \label{fig:interpolation-inr-without-noise:c}
    \end{subfigure}
    
    \caption{INR interpolation experiments without noise added.}
    \label{fig:interpolation-inr-without-noise}
\end{figure}

The rationale behind introducing a noise perturbation is that the very design of a ScaleGMN encoder ensures that functionally equivalent input networks differing only in a group action being applied will be mapped to the same point.
For completeness, we provide the results of performing the interpolation experiment without noise for the INR setting in Figure \ref{fig:interpolation-inr-without-noise}.
The observed behavior of the ScaleGMN autoencoder aligns with theoretical expectations, as all three transformations are mapped to an identical point in the latent space, yielding a flat interpolation.
This invariance to permutations is similarly an anticipated characteristic of Neural Graphs. Conversely, upon the introduction of scaling, there is no guarantee that both models will be encoded to the same latent point, a phenomenon illustrated in Figure~\ref{fig:umap}.
Nonetheless, it is empirically observed that the reconstructed models coincide in practice (Figures~\ref{fig:interpolation-inr-without-noise:b} and \ref{fig:interpolation-inr-without-noise:c}).
This ability is attributed to the learning process guided by the functional loss of the autoencoder architecture.
Despite the model weights being sign-swapped, the underlying functions they represent remain equivalent, and the autoencoder has consequently learned to assign a common canonical mapping to both.

\subsection{INR interpolation in latent space}
\label{appendix:interpolation-latent}

In Section~\ref{sec:exp}, we focused on weight space interpolation, as typically done in the model merging literature.
However, latent space interpolation is also common practice, as demonstrated by \citet{DeLuigi_2023}. 
Within the autoencoder framework, interpolation in weight space presents a notable computational advantage over interpolation in latent space. The latter method requires generating a sequence of $N$ latent points, each needing a separate forward pass through the decoder. Consequently, if the decoder's computational complexity were super-linear with respect to the number of input network parameters, this process would be markedly less efficient than weight-space interpolation, which exhibits linear complexity. However, as we use an MLP decoder, its decoding cost is also linear, thereby neutralizing this specific efficiency gain.

\definecolor{latent_scalegmn}{RGB}{203, 97, 32}
\definecolor{latent_ngraphs}{RGB}{129, 189, 129}

\begin{figure}[ht!]
    \centering
    \legendbox{%
    \legenditem{rebasin}{Linear Assignment}\hspace{1.5em}%
    \legenditem{latent_ngraphs}{Latent Neural Graphs}\hspace{1.5em}%
    \legenditem{latent_scalegmn}{Latent ScaleGMN}\hspace{1.5em}%
    }
    
    \vspace{4pt}

    \begin{subfigure}[b]{0.32\linewidth}
    \includegraphics[width=\linewidth]{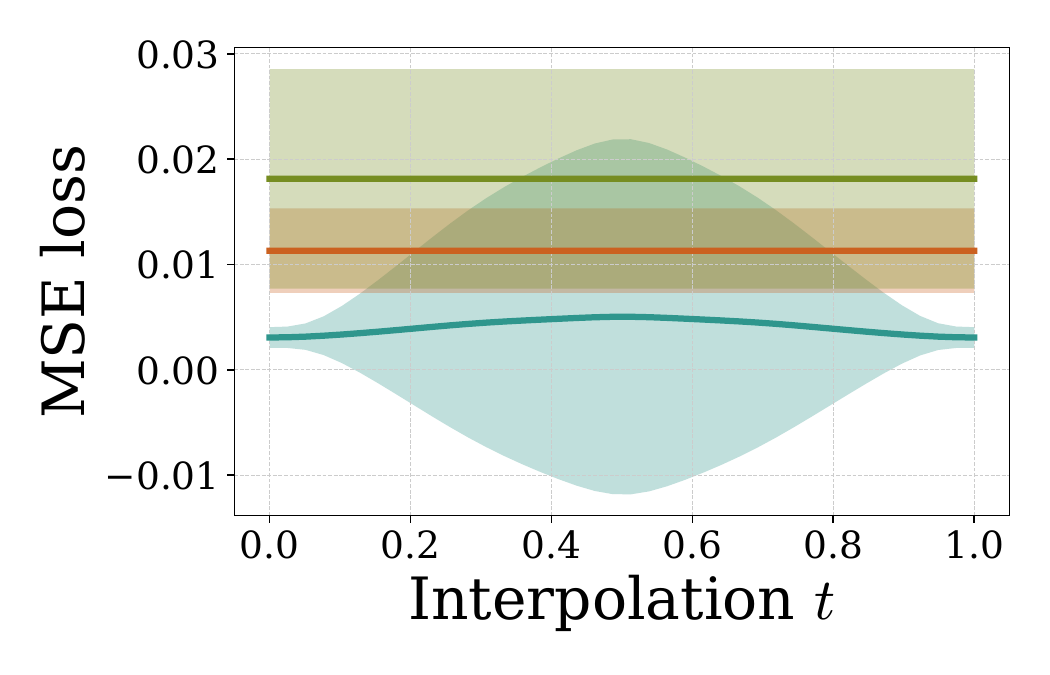}
    \caption{Permutations}
    \end{subfigure}\hfill
    \begin{subfigure}[b]{0.32\linewidth}
    \includegraphics[width=\linewidth]{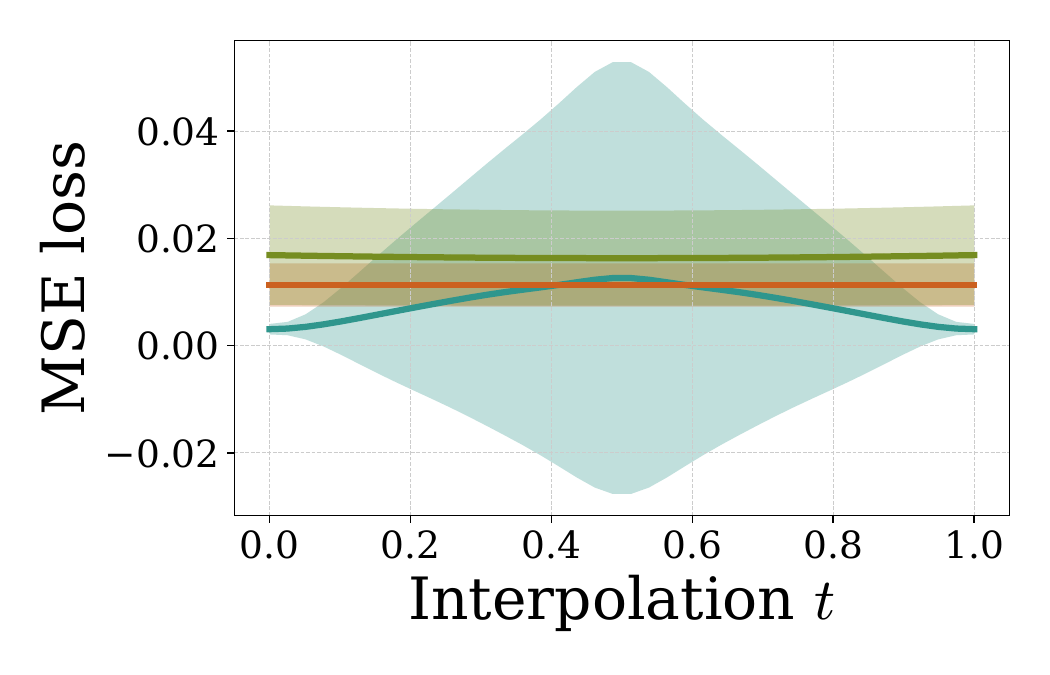}
    \caption{Scalings}
    \end{subfigure}\hfill
    \begin{subfigure}[b]{0.32\linewidth}
    \includegraphics[width=\linewidth]{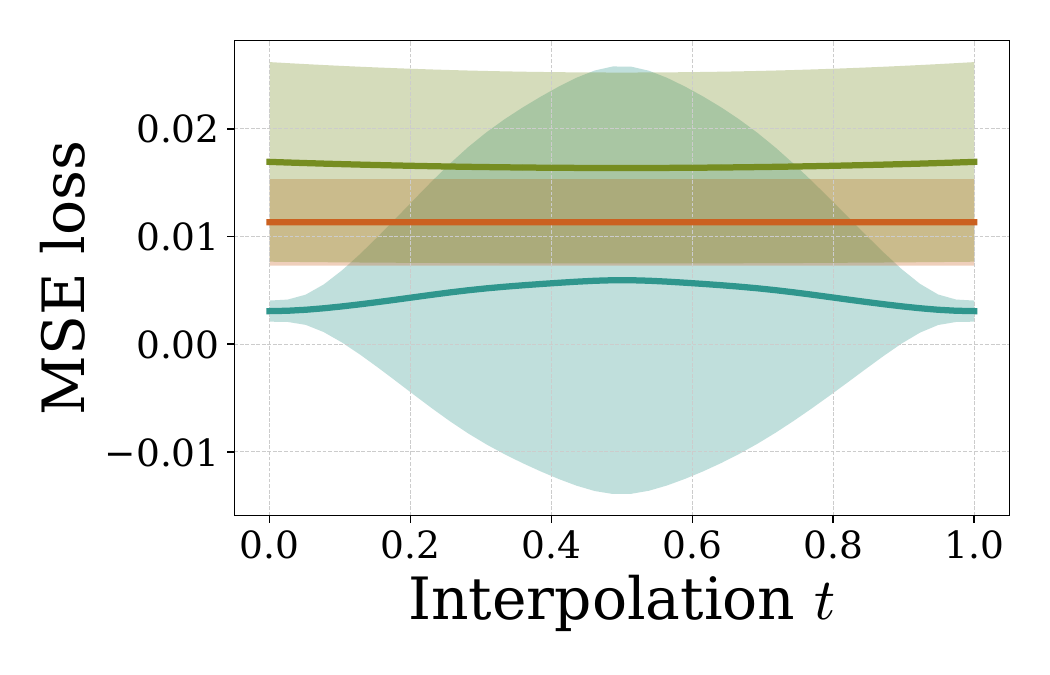}
    \caption{Permutations + Scaling}
    \end{subfigure}
    
    \begin{subfigure}[b]{0.32\linewidth}
    \includegraphics[width=\linewidth]{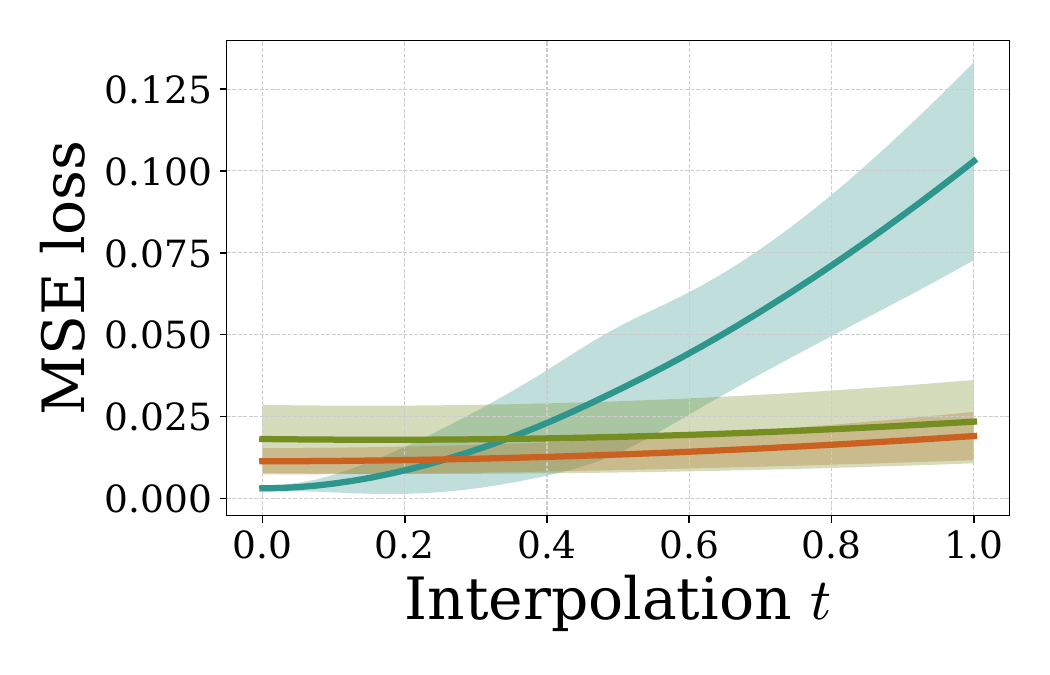}
    \caption{Permutations (Noise)}
    \end{subfigure}\hfill
    \begin{subfigure}[b]{0.32\linewidth}
    \includegraphics[width=\linewidth]{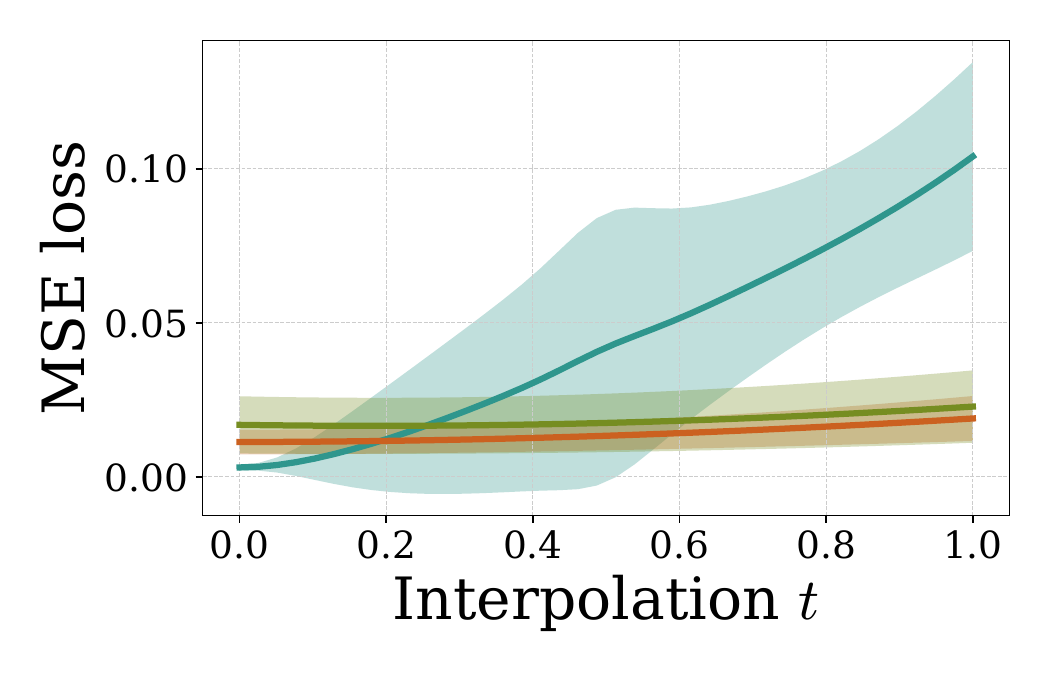}
    \caption{Scalings (Noise)}
    \end{subfigure}\hfill
    \begin{subfigure}[b]{0.32\linewidth}
    \includegraphics[width=\linewidth]{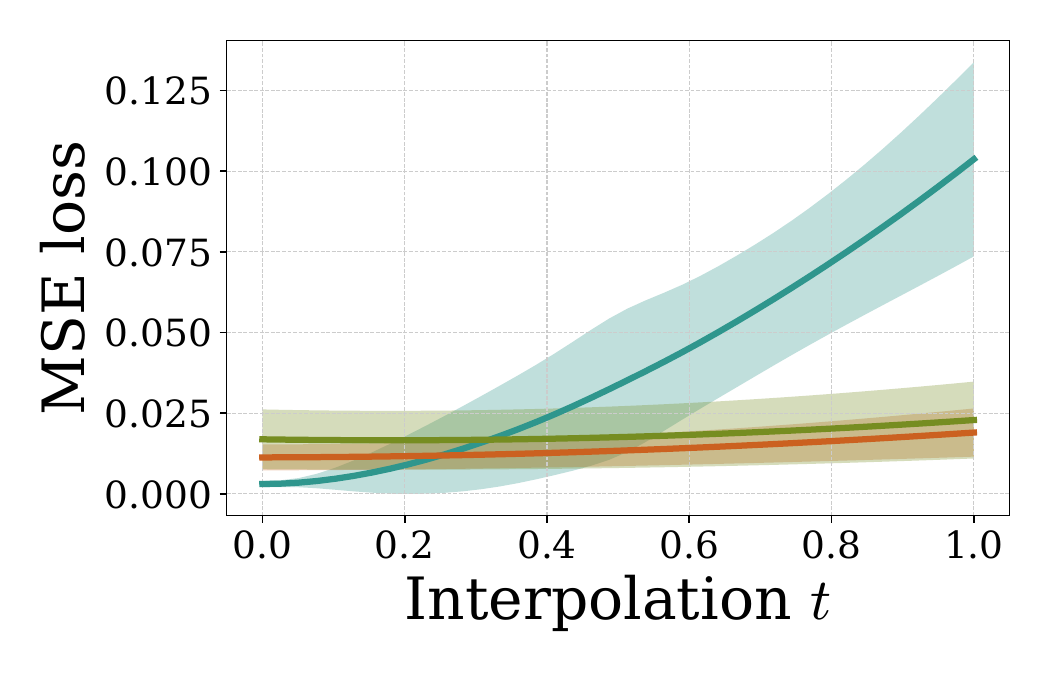}
    \caption{Permutations + Scaling (Noise)}
    \end{subfigure}
    
    \caption{INR latent interpolation experiments.}
    \label{fig:latent-interpolation-inr}
\end{figure}

We therefore investigate interpolation directly in the latent space produced by our autoencoders. As expected, Figure~\ref{fig:latent-interpolation-inr} shows that latent space interpolation on functionally identical INR pairs performs on par with our canonicalized weight space interpolation (Figure~\ref{fig:interpolation-inr}). This demonstrates that our learned latent representations are robust to perturbations in the encoded weights, which, as discussed in Section~\ref{sec:dis_imp_scale}, will prove valuable for downstream tasks beyond model merging.

\newpage
\subsection{CNN interpolation additional results}
\label{appendix:cnn-interpolation-additional}

\begin{figure}[th!]
    \centering

    % --- Define a variable for the subfigure width ---
    % 4 images per row, so width is slightly less than 1/4 = 0.25
    \newcommand{\subfigwidth}{0.24\linewidth}

    \vspace{1em}

    % Common Legend for both blocks
    \legendbox{%
      \legenditem{naivec}{Naive}\hspace{1.5em}%
      \legenditem{rebasin}{Linear Assignment}\hspace{1.5em}%
      \legenditem{ngraphs}{Neural Graphs}\hspace{1.5em}%
      \legenditem{scalegmn}{ScaleGMN}%
    }
    \vspace{1em}

    % --- FIRST BLOCK: Wrapped in a subfigure for a clean sub-caption ---
    \begin{subfigure}{\textwidth}
        \centering
        % Top row of the first block (Accuracies)
        % \includegraphics[width=\subfigwidth]{media/interpolation_cnn/interpolation_curves_accuracies_relu_7.pdf}\hfill
        \includegraphics[width=\subfigwidth]{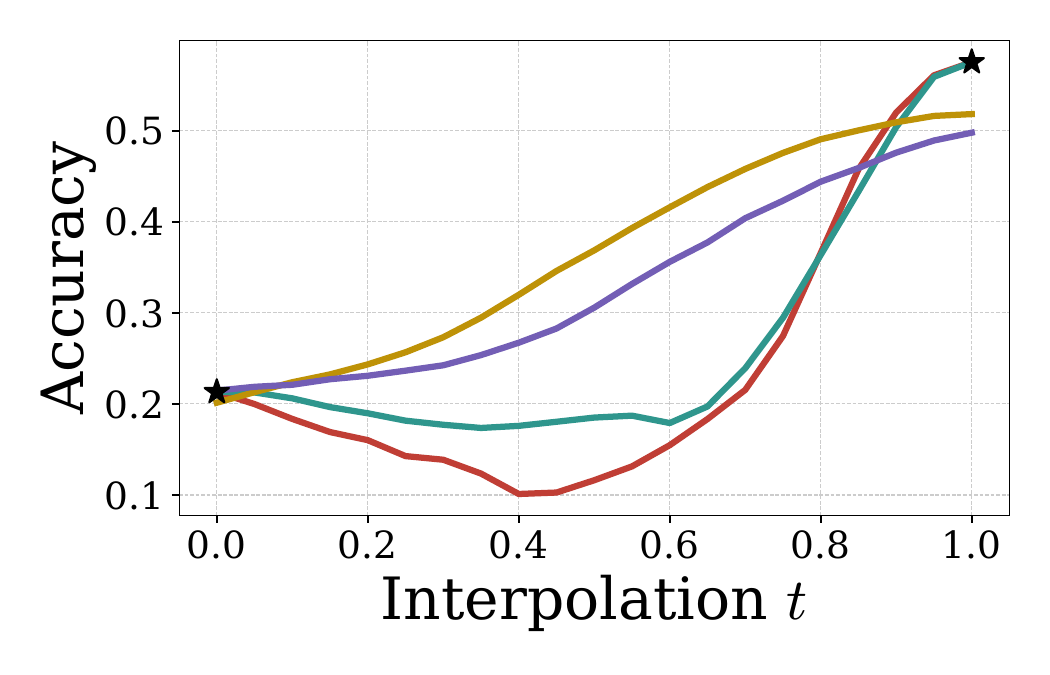}\hfill
        \includegraphics[width=\subfigwidth]{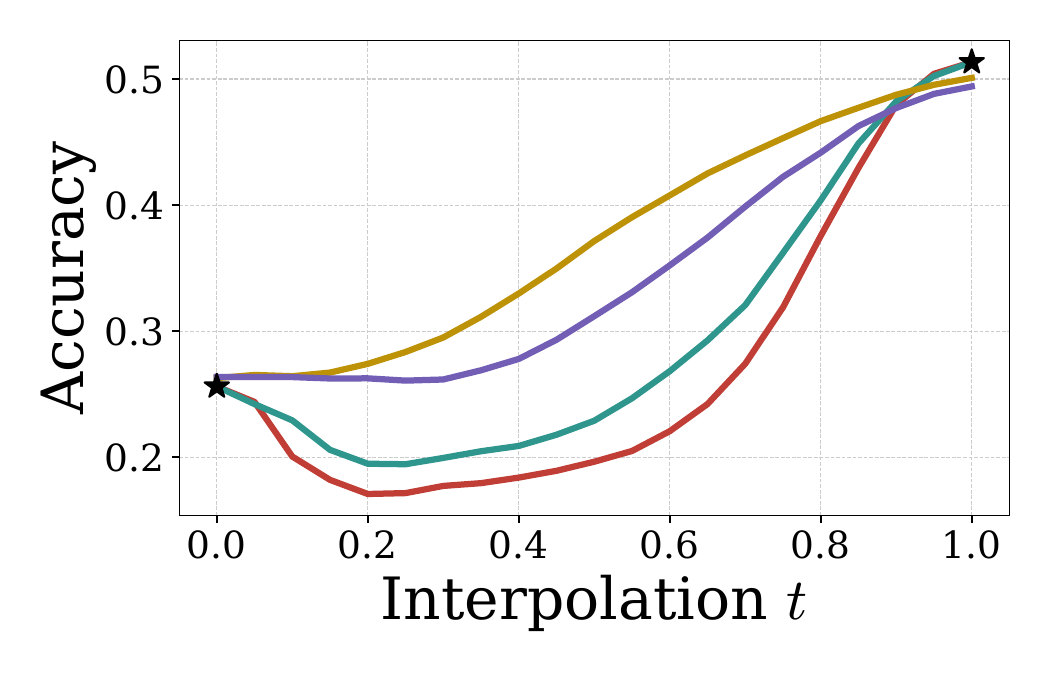}\hfill
        \includegraphics[width=\subfigwidth]{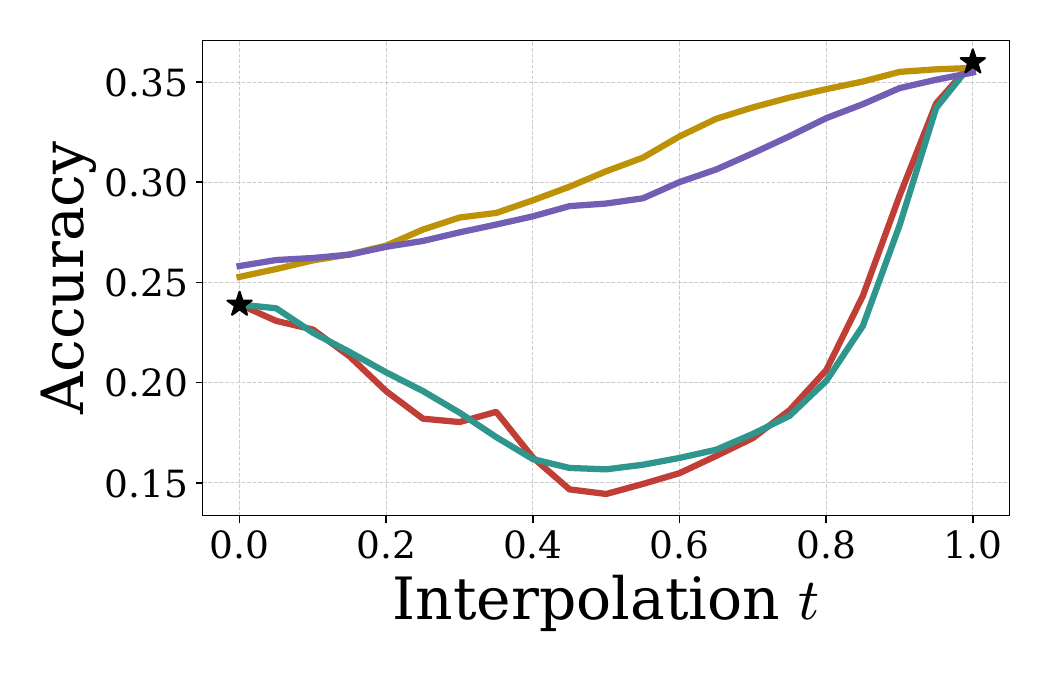}\hfill
        \includegraphics[width=\subfigwidth]{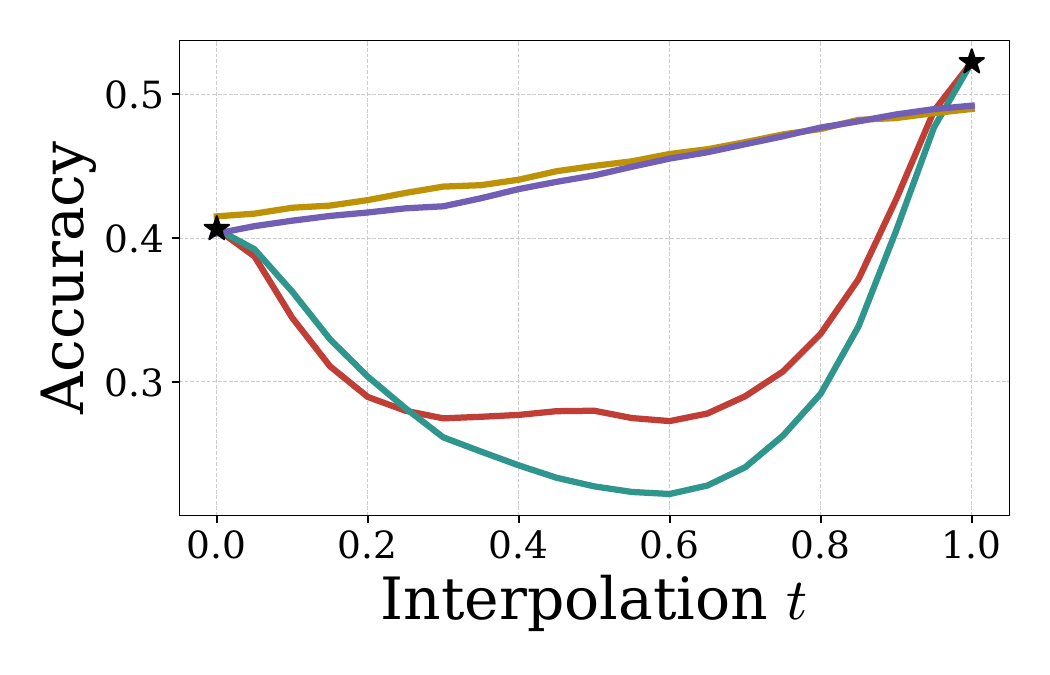}
        
        \vspace{2pt}

        % Bottom row of the first block (Losses)
        % \includegraphics[width=\subfigwidth]{media/interpolation_cnn/interpolation_curves_losses_relu_7.pdf}\hfill
        \includegraphics[width=\subfigwidth]{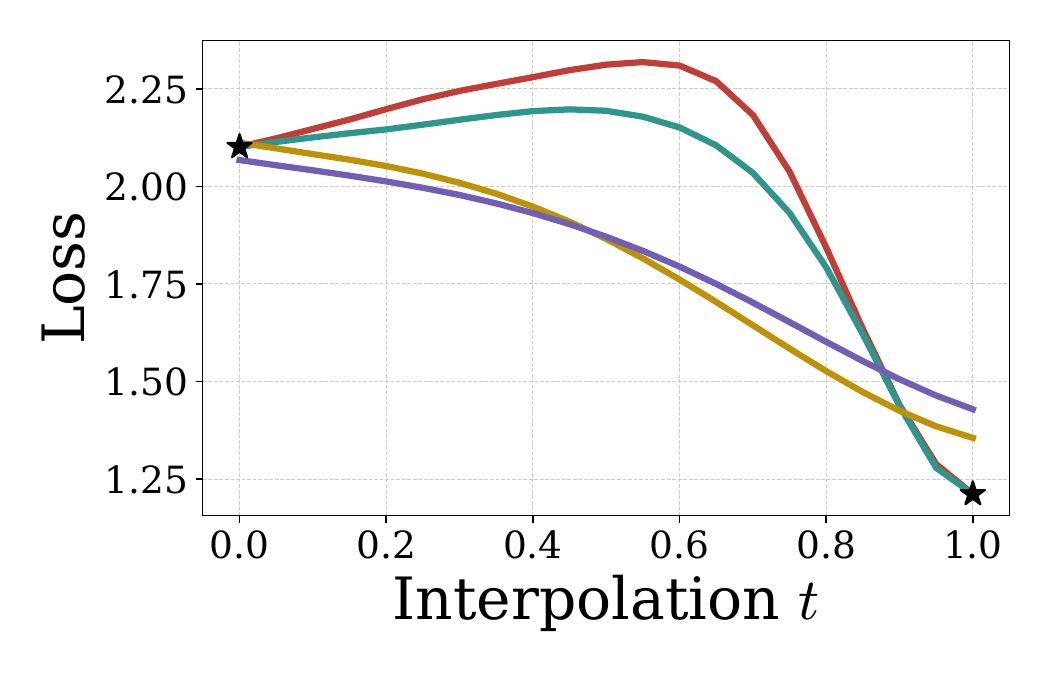}\hfill
        \includegraphics[width=\subfigwidth]{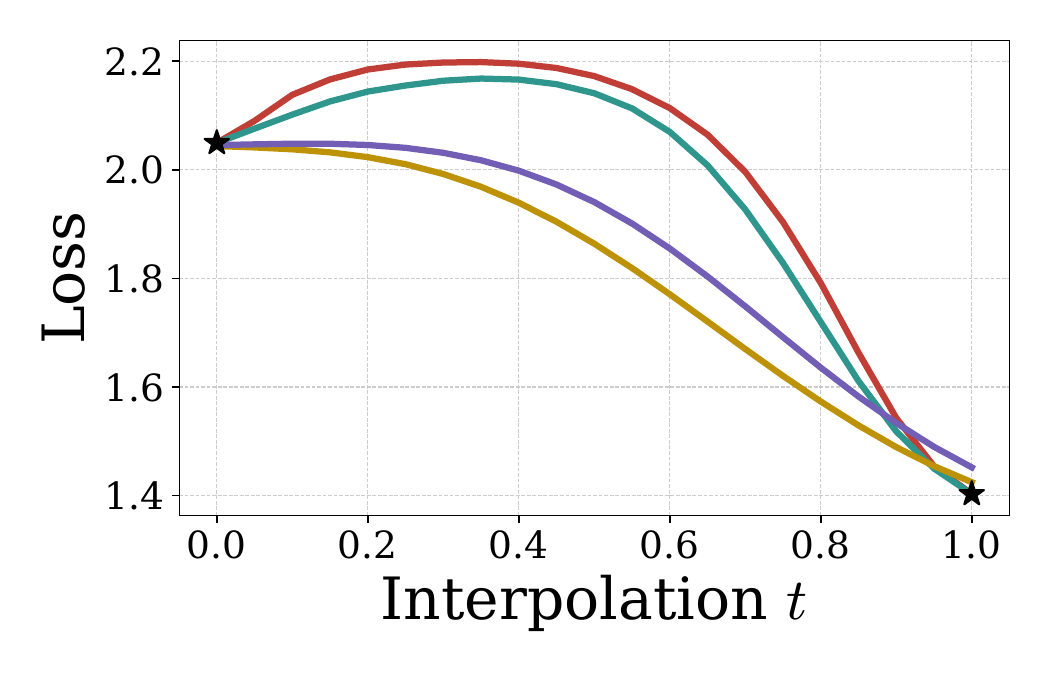}\hfill
        \includegraphics[width=\subfigwidth]{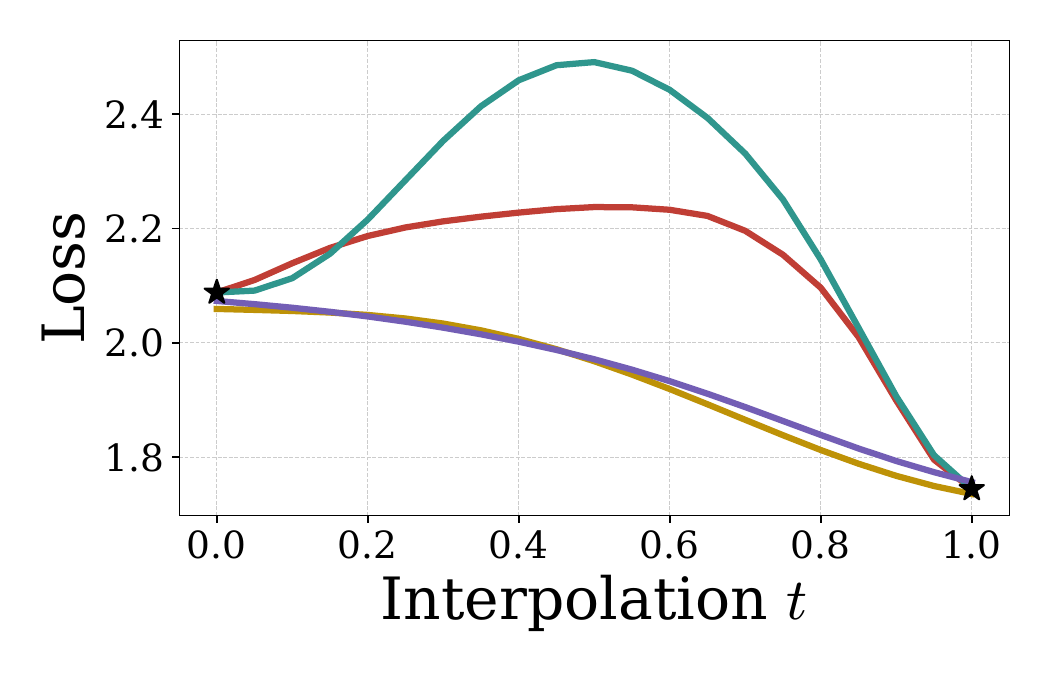}\hfill
        \includegraphics[width=\subfigwidth]{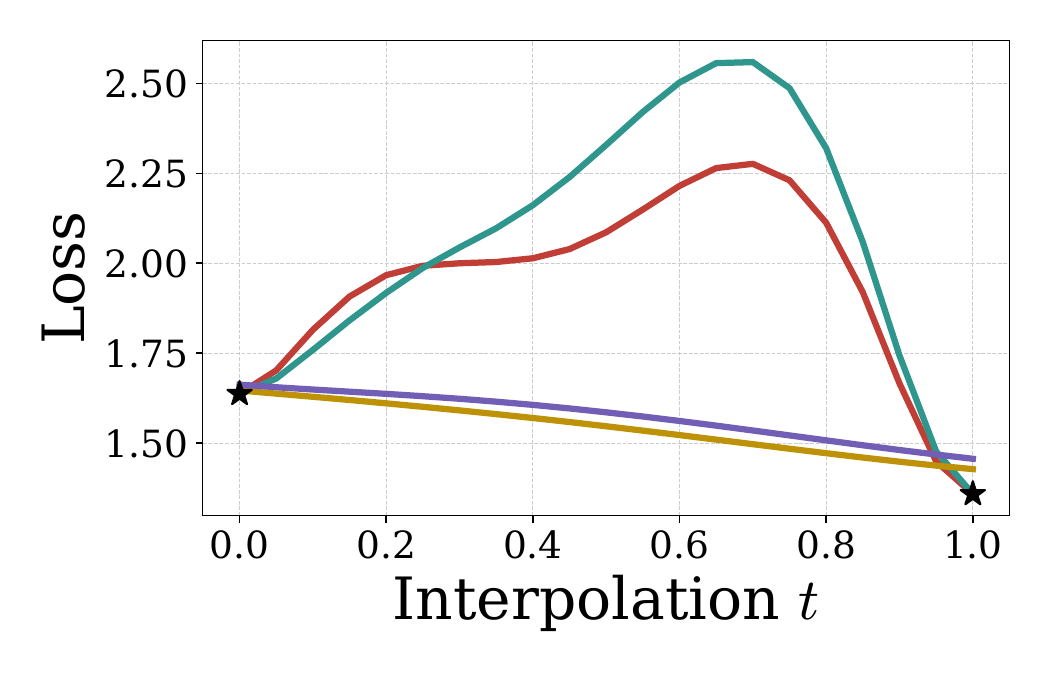}
        
        \caption{Models with ReLU activation.}
        \label{fig:interpolation-cnn-relu-extra}
    \end{subfigure}

    \vspace{2.5em} % Vertical space between the two blocks

    % --- SECOND BLOCK: Wrapped in a subfigure for a clean sub-caption ---
    \begin{subfigure}{\textwidth}
        \centering
        % Top row of the second block (Accuracies)
        % \includegraphics[width=\subfigwidth]{media/interpolation_cnn/interpolation_curves_accuracies_tanh_1500_500b.pdf}\hfill
        \includegraphics[width=\subfigwidth]{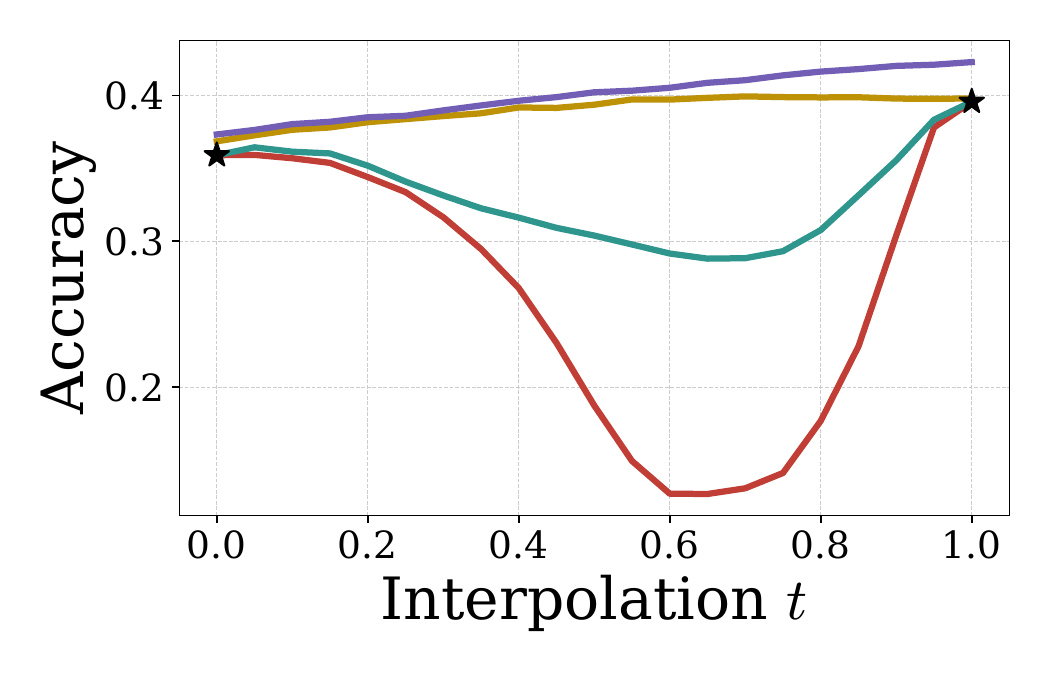}\hfill
        \includegraphics[width=\subfigwidth]{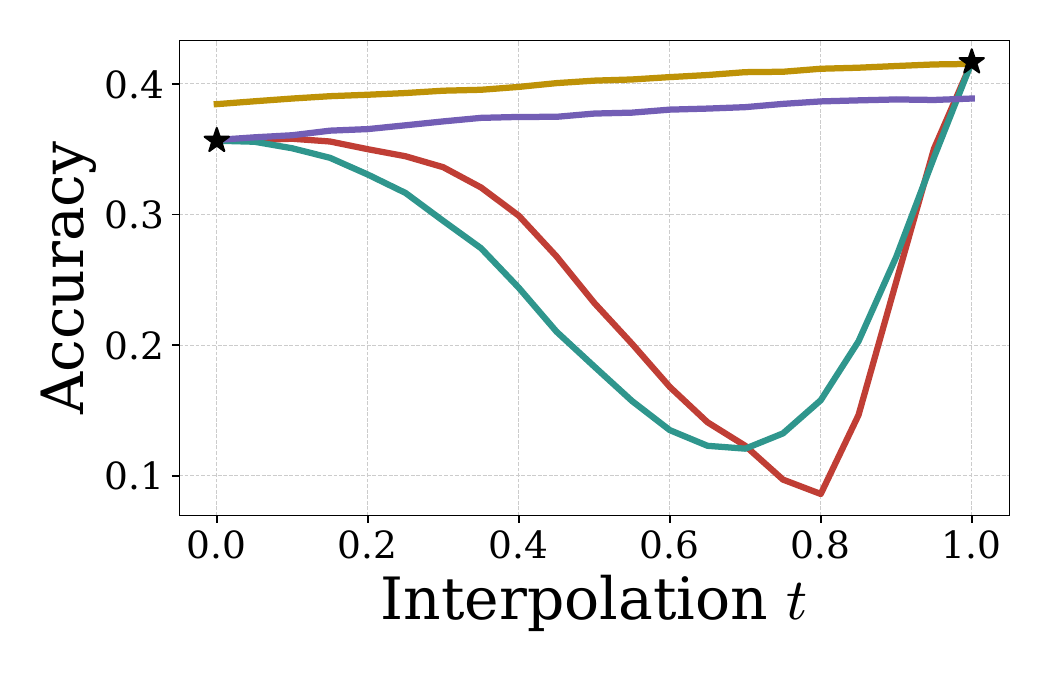}\hfill
        \includegraphics[width=\subfigwidth]{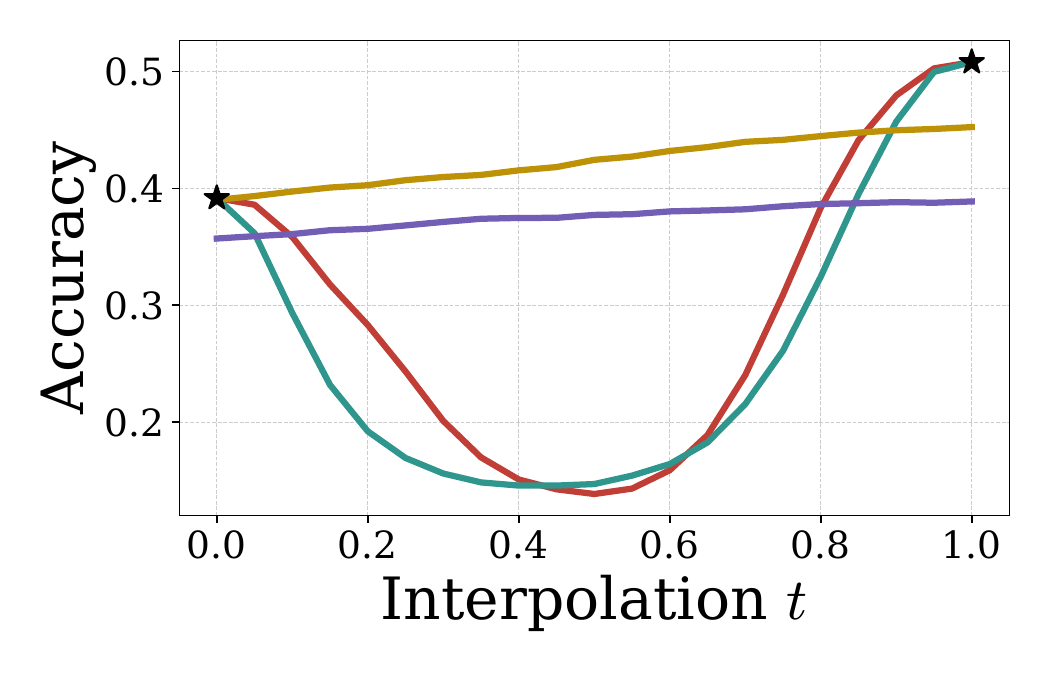}\hfill
        \includegraphics[width=\subfigwidth]{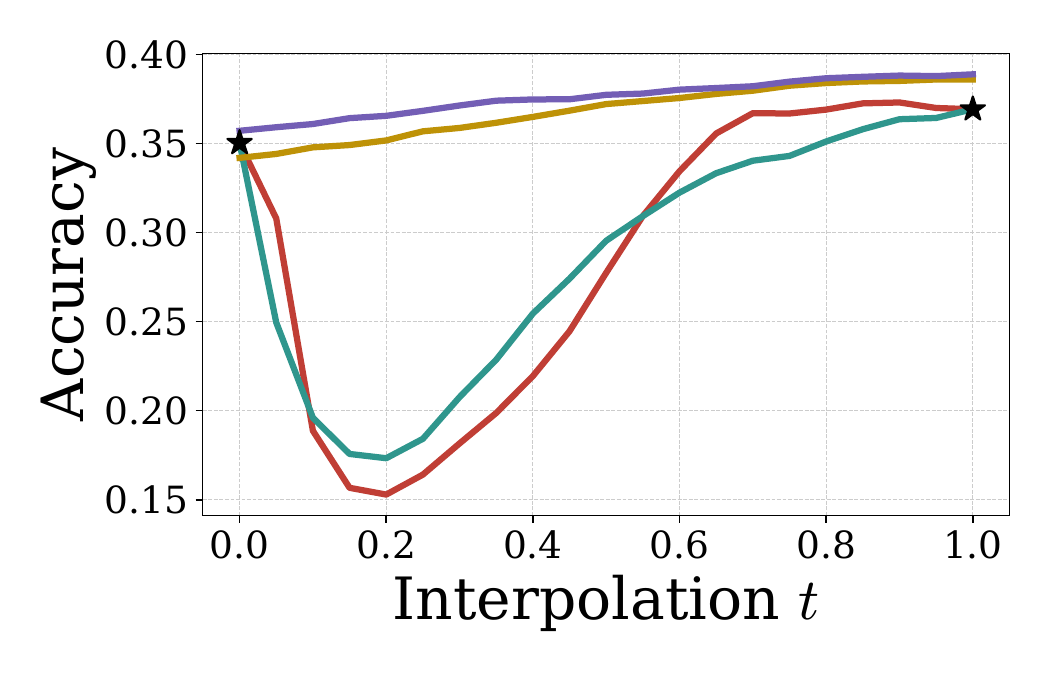}

        \vspace{2pt}
        
        % Bottom row of the second block (Losses)
        % \includegraphics[width=\subfigwidth]{media/interpolation_cnn/interpolation_curves_losses_tanh_1500_500b.pdf}\hfill
        \includegraphics[width=\subfigwidth]{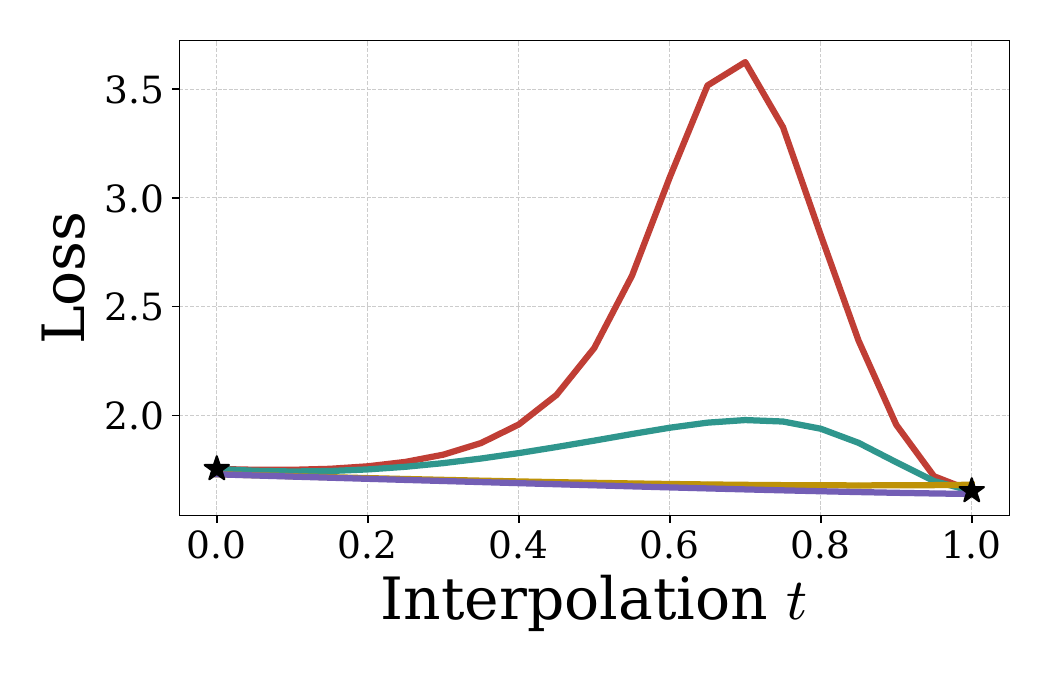}\hfill
        \includegraphics[width=\subfigwidth]{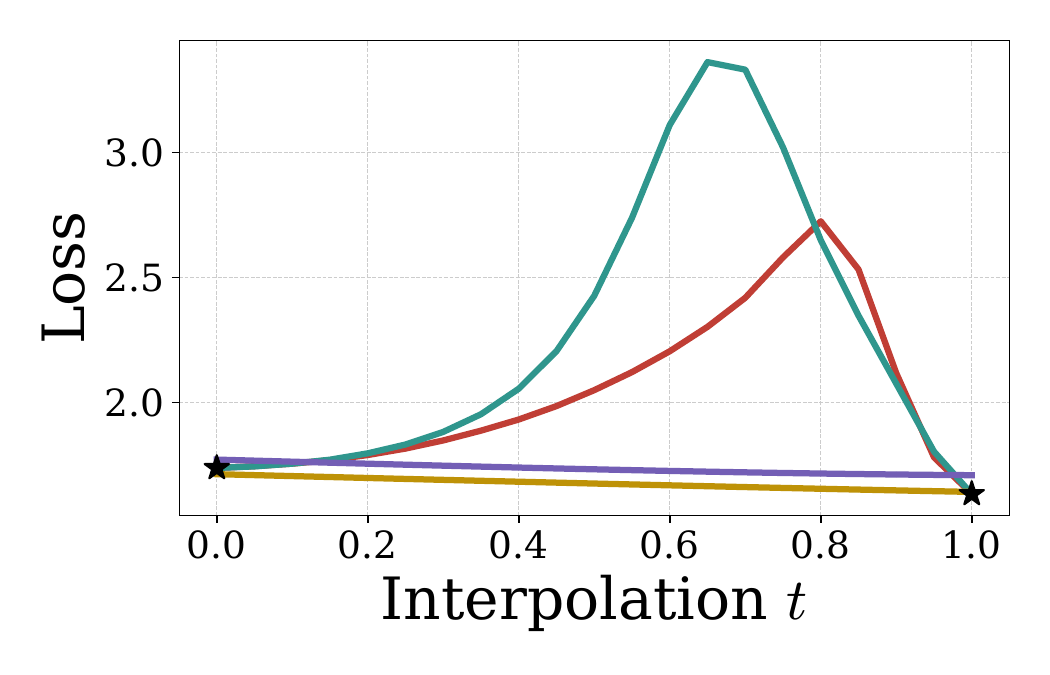}\hfill
        \includegraphics[width=\subfigwidth]{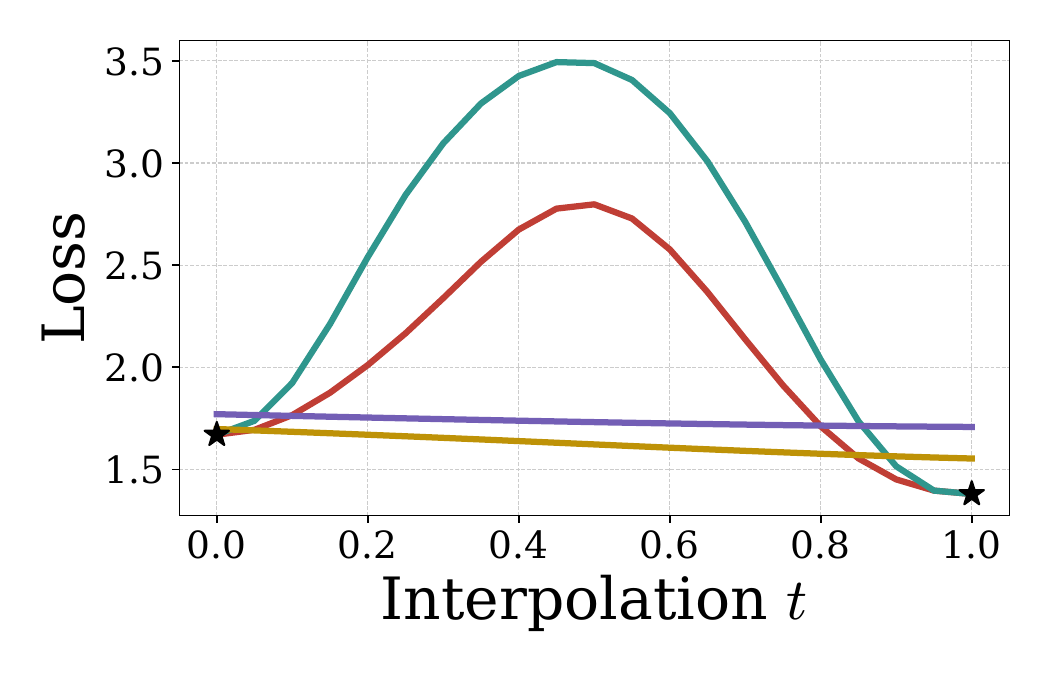}\hfill
        \includegraphics[width=\subfigwidth]{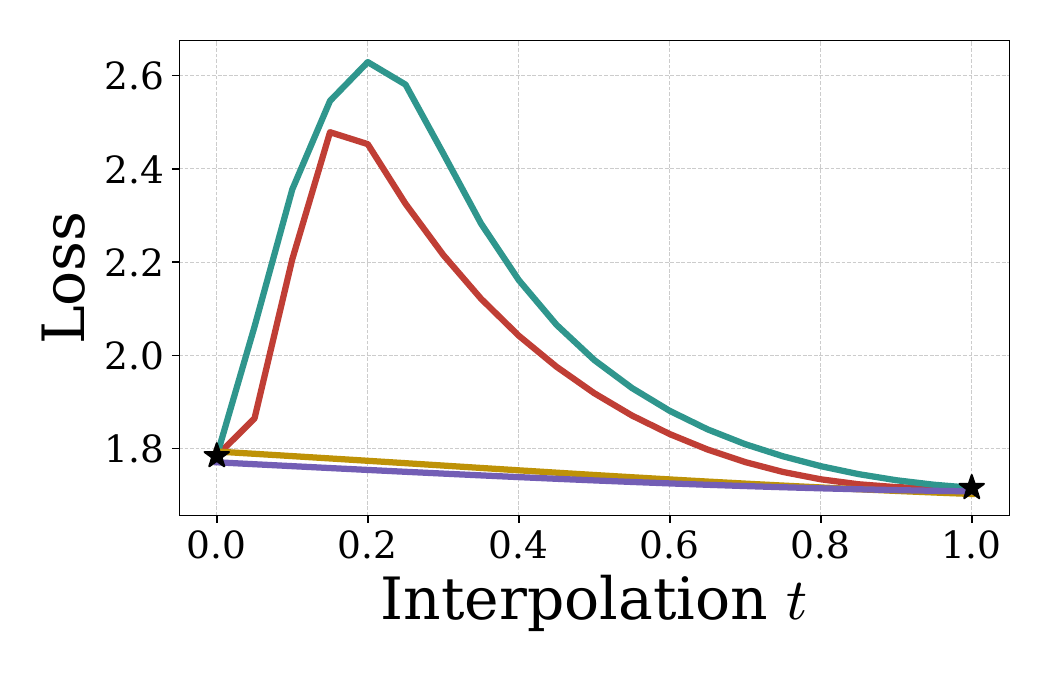}
        
        \caption{Models with Tanh activation.}
        \label{fig:interpolation-cnn-tanh-extra}
    \end{subfigure}
    
    \vspace{1em}
    
    \caption{Interpolation curves for CNN models comparing different activation functions across 4 model pairs each. For each block, the top row displays accuracies and the bottom row displays losses.}
    \label{fig:interpolation-cnn-combined}

\end{figure}

The interpolation results presented previously in Figures~\ref{fig:interpolation-cnn-relu-multiple-pairs} and \ref{fig:interpolation-cnn-tanh-multiple-pairs} were derived from Convolutional Neural Networks (CNNs) selected for their high accuracy, as these represent the most performant models and thus the most interesting for model merging in practical terms.
However, given the compact architecture, a significant portion of independently trained models from the SmallCNN Zoo dataset achieve suboptimal performance, with accuracies on the CIFAR dataset falling as low as 20\%.
In Figure \ref{fig:interpolation-cnn-combined} we present interpolation curves for randomly chosen model pairs to provide a more realistic and comprehensive view of the interpolation behavior across a wider spectrum of model quality.
In contrast to the Implicit Neural Representation (INR) experiments, the observed irregularities demonstrate that these CNNs constitute a more heterogeneous population of networks.

\end{document}